\documentclass[10pt]{article} 
\usepackage[accepted]{tmlr}

\usepackage{amsmath,amsfonts,bm}

\def\eqref#1{equation~\ref{#1}}

\def\1{\bm{1}}

\DeclareMathAlphabet{\mathsfit}{\encodingdefault}{\sfdefault}{m}{sl}
\SetMathAlphabet{\mathsfit}{bold}{\encodingdefault}{\sfdefault}{bx}{n}

\usepackage{graphicx} 
\usepackage{subfig} 
\usepackage{hyperref}
\usepackage{url}
\usepackage[bb=dsserif]{mathalpha} 
\usepackage[group-separator={,}]{siunitx} 
\usepackage{algorithm}
\usepackage{etoolbox}
\let\classAND\AND
\let\AND\relax
\usepackage{algorithmic}

\let\AND\classAND
\AtBeginEnvironment{algorithmic}{\let\AND\algoAND}
\def\HH{H\!H}

\title{Hidden Heterogeneity: When to Choose Similarity-Based Calibration}

\author{\name Kiri L. Wagstaff \email kiri.wagstaff@oregonstate.edu \\
      \addr School of Electrical Engineering and Computer Science\\
      Oregon State University
      \AND
      \name Thomas G. Dietterich \email tgd@cs.orst.edu \\
      \addr School of Electrical Engineering and Computer Science\\
      Oregon State University
}

\usepackage{changes}
\renewcommand{\comment}[1]{}

\def\month{1}  
\def\year{2023} 
\def\openreview{\url{https://openreview.net/forum?id=RA0TDqt3hC}} 

\begin{document}

\maketitle

\begin{abstract}
Trustworthy classifiers are essential to the adoption of machine
learning predictions in many real-world settings.  The predicted
probability of possible outcomes can inform high-stakes decision
making, particularly when assessing the expected value of alternative
decisions or the risk of bad outcomes.  These decisions require
well-calibrated probabilities, not just the correct prediction of the
most likely class.  Black-box
classifier calibration methods can \textit{improve the reliability} of a
classifier's output without requiring retraining.  However, these
methods are unable to detect subpopulations where calibration could also
\textit{improve prediction accuracy}. Such subpopulations are said to
exhibit ``hidden heterogeneity'' (HH), because the original classifier
did not detect them. This paper proposes a quantitative measure for HH.
It also introduces two similarity-weighted calibration methods that
can address HH by adapting 
locally to each test item: SWC weights the calibration set by
similarity to the test item, and SWC-HH explicitly incorporates hidden
heterogeneity to filter the calibration set.
Experiments show that the improvements in calibration
achieved by similarity-based calibration methods correlate with the
amount of HH present and, given sufficient calibration data, generally
exceed calibration achieved by global methods. 
HH can therefore serve as a useful diagnostic
tool for identifying when local calibration methods would be beneficial.
\end{abstract}

\section{Introduction}\label{sec1}

How do we know when to trust a prediction?  Let $f(X)$ be a classifier
that outputs a discrete probability distribution $\hat{P}(Y|X)$ over
the $K$ possible class labels $\{1,\ldots,K\}$.
Ideally, each prediction made by the classifier will be {\em
point-wise calibrated}, that is, the true class distribution for each $X$
matches $\hat{P}$: $P(Y|X) = \hat{P}(Y|X)$.  Many investigators
have studied a weaker requirement that each distinct output value
$g$ is calibrated, such that $P(Y|g) = E[Y]$ over the set of
predictions for which $\hat{P}(Y|X) = g$ ~\citep{vaicenavicius:calib19,widmann:calib19}. 
This ensures aggregate calibration for the set, but it allows
individual predictions to have $P(Y|X) \neq \hat{P}(Y|X)$.
Others use the same approach but require calibration only for the
most-likely class's predicted
probability~\citep[e.g.,][]{guo:calib17,patel:mi-calib21,luo:local22}
or marginal class
probabilities~\citep[e.g.,][]{zadrozny:iso02,kumar:calib19}, both of
which also do not enforce point-wise calibration. 
Point-wise calibration of each individual prediction, for its full class
distribution, is important for safety-critical applications.
Well-calibrated predictions improve the trustworthiness of systems and
support downstream cost-sensitive decisions (e.g.,~medical diagnosis,
autonomous driving, financial decisions).
Likewise, calibration is necessary when combining or comparing
predictions from different sources~\citep{bella:sba13} or in
classifier cascades that use a low-cost but less accurate classifier's
output to decide whether to apply a higher-cost but more accurate
secondary classifier~\citep{enomoto:cascade21}.  Good calibration is
beneficial in any decision making setting in which uncertainty matters
(e.g., active learning or classification with a rejection or
abstention option).

We focus on an increasingly common use case in which we would like to
apply a pre-trained, possibly proprietary, model $\mathcal{M}$ to our
own data set $\mathcal{D}$ with corresponding distribution $P_D(Y
\vert X)$.  In this scenario, the original training data set is
unavailable and $P(Y \vert X)$, the distribution for which
$\mathcal{M}$ was trained (and possibly calibrated), is unknown.  Any
domain shift between $P$ and $P_D$ could prevent 
$\mathcal{M}$ from generating reliable predictions on $\mathcal{D}$.
Moreover, even in the absence of domain shift,
$\mathcal{M}$ may perform poorly on $D$ due to ``hidden
heterogeneity'', which occurs when $\mathcal{M}$ assigns the same
posterior probability to items with different true probabilities.


Concerns about poorly calibrated classifiers are not
new~\citep[e.g.,][]{zadrozny:hist01,niculescu-mizil:calib05}, and
several post-training calibration correction
methods have been
developed~\citep[e.g.,][]{guo:calib17,kumar:calib19,kull:dirichlet-calib19,alexandari:bcts20}. 
In general, these methods devise a {\em calibration map} $\Phi$ that
transforms the original predicted probabilities into values that are
better calibrated.  We denote the output of a
classifier $f(x_i)$ applied to item $x_i$ as the probability vector
$\hat{p}_i$ of length $K$ (number of classes) that sums to 1
(i.e.,~resides in the simplex $\Delta_{K-1}$).  The calibration map
$\Phi: \Delta_{K-1} \mapsto \Delta_{K-1}$ is derived from an
independent calibration set $\mathcal{C}$ to transform $\hat{p_i}$ to
a more reliable $\hat{q}_i = \Phi(\hat{p}_i)$.

A key limitation of these calibration maps is the implicit assumption
that all items with the same predicted probability vector $\hat{p}$
should be given the same correction.  Such maps cannot accommodate hidden
heterogeneity, which manifests as 
subpopulations with distinct $P(Y|X)$ values to
which the classifier has erroneously assigned 
the same $\hat{p}$ value.  For example, in predicting cancer risk,
there could be many different reasons (age, lifestyle, family medical
history, etc.)~that a given individual is predicted to have $\hat{p} =
0.9$.  Even if the predictions satisfy aggregate calibration,
this probability could be an over-estimate for some individuals, while for 
others it could be an under-estimate.  No global calibration map
can address this heterogeneity to achieve point-wise calibration,
because they map all items with the same $\hat{p}$ to the same $\hat{q}$.


We propose a method to quantify hidden heterogeneity (HH) as a signal
for when global calibration may be inadequate.
%
Once HH is detected, we face a choice of either (a) training a new
classifier on the available calibration data or (b) improving the
existing classifier using the calibration data. Because HH is a local
phenomenon, a natural way to improve the classifier is to apply a
local, similarity-based calibration technique.
We introduce two local calibration methods that 
leverage the location of $x_i$ in feature
space to yield $\hat{q}_i = \Phi(\hat{p}_i \vert x_i)$.  These methods
determine the calibrated probability $\hat{q}_i$ by taking a weighted
vote of data points in the calibration set $\mathcal{C}$.
The first method, Similarity-Weighted Calibration (SWC), assigns
weights to every point in the calibration set based on similarity to
$x_i$.
The second
method, SWC-HH, uses only items within a local neighborhood defined by
the estimated HH.  We refer to the weighted number of calibration data
points as the ``calibration support'' for $x_i$,  which indicates how
much calibration data is available for estimating $\hat{q}_i$.
This measure of the calibration quality of $\hat{q}_i$ for each $x_i$ is
a unique advantage of local calibration.

We note that any post-hoc calibration method can be viewed as a form
of model stacking~\citep{wolpert:stacking92}, in which the output of
the original classifier is transformed via $\Phi$, a model itself.
Our SWC and SWC-HH methods are stacking methods that focus on
improving local calibration.  As a consequence, they also reduce or
eliminate HH and can thereby improve classifier accuracy.

The major contributions of this paper are %
\begin{enumerate}
  \vspace{-.15in}
  \item The identification of hidden heterogeneity as a property of
    classifier predictions that thwarts global calibration methods
    (Section~\ref{sec:hh}),%
  \vspace{-.1in}
  \item A method for quantifying hidden heterogeneity to indicate when
    local calibration is needed (Section~\ref{sec:hh}),%
  \vspace{-.1in}
  \item Two local calibration methods based on Similarity-Weighted Calibration
    (Section~\ref{sec:swc}), and %
  \vspace{-.1in}
  \item Results of experiments that assess the relationship between
    hidden heterogeneity and calibration, yielding useful guidance for
    practitioners (Section~\ref{sec:res}).
  \vspace{-.1in}
\end{enumerate}
We provide context from previous work in Section~\ref{sec:relwork}.
Key conclusions and limitations of local, similarity-based calibration are
discussed in Section~\ref{sec:conc}. 

\section{Related Work}
\label{sec:relwork}

There are several methods for improving the reliability (calibration)
of classifier predictions.  Many recent advances were inspired
by the recognition that deep neural networks in particular may
sacrifice calibration to achieve higher generalization
accuracy~\citep{guo:calib17}.  Strategies include using
calibration-sensitive training methods, if the original training set
is available~\citep[e.g., via modifications to the loss function, as
  proposed
  by][]{kumar:mmce18,mukhoti:focal-loss20,enomoto:cascade21,tomani:adversarial21},
using domain-specific representations that lead to improved
calibration~\citep{kalmady:calibHF21}, or adopting network
architectures that do not use convolutions~\citep{minderer:calib21}.
In contrast, post-hoc calibration correction methods that directly modify the
classifier's predictions on new observations, without re-training, can
be employed even when the training data (or model) are proprietary or
when the data distribution has changed and we wish to recalibrate an
existing classifier to extend its applicability.

Global, parametric calibration methods re-map the predicted
probabilities, $\hat{p}$, output by the classifier by fitting a chosen
functional form (e.g.,~logistic curve) from the probabilities to the
labels to compute $\hat{q} = \Phi(\hat{p})$.  For binary
classifiers, Platt scaling~\citep{platt:calib99} transforms $\hat{p}_i$ into
a value between 0 and 1 using a sigmoid function with two parameters,
$A$ and $B$: $\hat{q}_i = \frac{1}{1 + e^{A \hat{p}_i + B}}$.  The
parameters $A$ and $B$ are chosen to optimize the negative
log-likelihood of predictions made on the calibration set.  Platt
scaling was generalized to multi-class problems for neural
networks~\citep{guo:calib17} via a method called temperature scaling,
which operates on the logits $z_i$ (not the probabilities) by
optimizing a temperature parameter $T$ in $u_i[k] = e^{z_i[k] / T}$, where
$z_i[k]$ is the logit for item $i$ and class $k$, and $u_i[k]$ is the
corresponding unnormalized probability.  These values are normalized
as $\hat{q}_i[k] = \frac{u_i[k]}{\sum_{j} u_j[k]}$.  The same $T$ value
is used for all classes.  Bias-Corrected Temperature
Scaling~\citep{alexandari:bcts20} adds a bias term for each class.

There are also several approaches that construct probability bins and
assign the average (or other
aggregate) accuracy within bin $B_b$ as its calibrated
probability, $\hat{q}_i := Acc_b, \forall i \in B_b$.  Histogram
binning~\citep{zadrozny:hist01} assigns items to bins based on their
uncalibrated predictions $\hat{p}_i$, often using equally-spaced bin
boundaries or divided so that each bin has the same number of items
(``equal frequency'') from the calibration set.  \cite{kumar:calib19} found
that the latter strategy, as well as using a larger number of bins,
yields better results.  Isotonic regression~\citep{zadrozny:iso02}
adds further flexibility by optimizing the bin boundaries to
minimize the squared loss between $\hat{q}_i$ and $y_i$.
Recently,~\cite{patel:mi-calib21} proposed selecting the bin
boundaries to maximize the mutual information between bin predictions
$\hat{q}_i$ and $y_i$. 

To date, very few calibration methods have leveraged the location of
items in feature space, $\mathcal{X}$.
%
\cite{zhao:indiv20} introduced ``individual'' (per-item) calibration for
regression problems and confidence intervals.
Partial specialization for classification problems can be achieved by
estimating a different $T$ per subpopulation (unlabeled
cluster~\citep{gong:multidom-ts21} or labeled
``domain''~\citep{yu:multidom-ts22}),
then employing linear regression to estimate a new $T'$ for each
test item.  Our approach operates at a finer (per-item) granularity
and is not restricted to probability rescaling.
Like our method, the LoRe calibration method~\citep{luo:local22}
considers the similarity of the calibration
set items to the test item $x$. However, LoRe restricts the similarity
calculation to calibration items that fall into a probability bin
based on the probability $\max_k \hat{p}_i[k]$ of the
highest-probability class. This can produce high variance estimates
when the bin contains few calibration items. Our method avoids this
problem by considering the full predicted distribution $\hat{p}_i$
when computing similarity. LoRe also only calibrates the
highest-probability prediction; it does not produce a calibrated
probability distribution over all $K$ classes. Consequently, it does
not support downstream tasks such as computing the expected costs of
misclassification (in cost-sensitive problems) or re-estimating class
probabilities~\citep{alexandari:bcts20}.

One calibration approach that employs similarity to compute the complete
$\hat{q}_i[k]$ vector is Similarity-Binning Averaging or
SBA-10~\citep{bella:sba09}, 
which creates bins (neighborhoods) that contain an item's 10 nearest
neighbors (in Euclidean distance) in an ``augmented''
feature space $\mathcal{X}^+ = 
\mathcal{X}\times \Delta_{K-1}$ defined by the item's feature 
vector $x_i$ of dimension $d$ concatenated with its probability
vector $\hat{p}_i \in \Delta_{K-1}$.  SBA-10
computes the calibrated probability $\hat{q}_i[k]$ as the 
probability of class $k$ (in the calibration set) within item $i$'s
assigned bin, with each 
item contributing equally~\citep{bella:sba09}.  In contrast, our approach
uses a similarity-weighted contribution from every item in the
calibration set, not just the 10 nearest neighbors.
%
%
%

\section{Hidden Heterogeneity}
\label{sec:hh}

Global post-hoc calibration methods, such as Platt scaling and
temperature scaling, perform very well for some data sets and
algorithms and less well for others. Similarly, local methods like SBA-10 do not
always improve upon these global methods. What causes the failure of
global methods, and under what conditions can local methods do better?
Our hypothesis is that global post-hoc calibration, which focuses on
aggregate calibration, fails when the 
data exhibits \textit{hidden heterogeneity} (HH)
with respect to the predicted probabilities $\hat{p}$. HH
characterizes situations where
there are subpopulations in the feature space $\mathcal{X}$
to which the classifier assigns 
the same $\hat{p}$ but that require different calibration corrections.

\subsection{Hidden Heterogeneity}

\paragraph{Definition 1}
A classifier $f$ exhibits {\em hidden heterogeneity} with respect to a
feature space $\mathcal{X}$ if there exists a subregion $\mathcal{U}
\subseteq \mathcal{X}$ such that $f(x) \approx \hat{p}$ for all $x \in
\mathcal{U}$ and yet $\mathcal{U}$ can be partitioned into $M$
disjoint subregions $\mathcal{U} = \mathcal{U}_1 \coprod \cdots
\coprod \mathcal{U}_M$ such that the true class probabilities
$P(y|x\in \mathcal{U}_m) \neq P(y|x \in \mathcal{U}_{m'})$ for all
distinct pairs $m,m' \in \{1,\ldots,M\}, m \neq m'$. 

An extreme example of HH occurs for a classifier that ignores all
features and predicts the majority class for all items.  Imagine a
data set composed of 60\% cats and 40\% birds, for which a classifier
predicts $P(y =$ ``cat''$) = \hat{p} = 0.6$ for all items (i.e.,
$\mathcal{U} = \mathcal{X})$.  This classifier is perfectly calibrated
in terms of aggregate calibration, but it is uninformative about any
individual animal.  If cats and birds are not separable in 
the feature space, this may be the best one can do.  However, 
if the items have a feature such as ``number of legs'', then
two subregions---$\mathcal{U}_1$ for animals with two legs and
$\mathcal{U}_2$ for animals with four legs---can be defined with true
conditional probabilities of 1 (for ``cats'') and 0 (for
``birds'').  This heterogeneity is hidden in the classifier's
predictions.

This extreme situation (complete HH) could happen for a number of
reasons (majority-class classifier, classifier only trained on cats,
etc.).  More commonly, any classifier 
may have one or more subregions $\mathcal{U}$ in its predicted
probabilities that likewise obscure informative heterogeneity,
due to model misspecification, an overly constrained hypothesis space,
over-regularization, or data shift.
Detecting HH can alert the practitioner to limitations of the
classifier.  While global methods that map $\hat{p}$ to
$\hat{q}$ cannot address HH, local calibration could model the
subregions separately and assign $\hat{q}_i$ differently for each
$\mathcal{U}_i$.

\comment{
The extent to which hidden heterogeneity can be detected is determined
by the representation of the input space $\mathcal{X}$.  
Consider a domain in which items from multiple classes are
not separable given their feature space representation. In this case,
the uncertainty in the class label is aleatory and
irreducible. However, if the input representation makes it possible to
separate the classes (e.g., based on number of legs), then the
uncertainty is epistemic (model-specific) and could be detected by a
different model.

There are many reasons why a learned model might exhibit hidden
heterogeneity. First, the training data
set might not have been large or diverse enough to reveal the
heterogeneity. Second, global regularization may have forced the
learned function $f$ to be too smooth in some regions of the input
space. A third case arises when the test distribution is different
than the training distribution. The learned function could have no
hidden heterogeneity with respect to the training distribution and yet
show hidden heterogeneity on the test set.
}

\begin{algorithm}[tb]
\caption{Hidden Heterogeneity (HH)}
\label{alg:hh}
\textbf{Input}: Test item $x_t$, calibration data $\mathcal{C}$,
predicted probabilities $\hat{p}$, and radius $r$ \\ 
\textbf{Output}: Hidden heterogeneity in neighborhood around $x_t$
\begin{algorithmic}[1] 
  \STATE Construct probability neighborhood around $x_t$:
  $\mathcal{U}_t = \{ x_i \in \mathcal{C} | D_H(\hat{p}_t,
  \hat{p}_i) < r \}$ (using Eqn.~\ref{eqn:hell}).
  \STATE Train $g_t$ using labeled data in $\mathcal{U}_t$.
  \STATE Collect model predictions for the neighborhood:
  $f(\mathcal{U}_t) = \{ \hat{p}_i | x_i \in \mathcal{U}_t \}$.
  \STATE Collect $g_t$ predictions for the neighborhood:
  $g_t(\mathcal{U}_t) = \{ g_t(x_i) | x_i \in \mathcal{U}_t \}$.
  \STATE Collect labels for the neighborhood: $Y_{\mathcal{U}_t} =
    \{ y_i | x_i \in \mathcal{U}_t \}$.
  \STATE Calculate $\HH_{\mathcal{U}_t}$ using $f(\mathcal{U}_t)$,
  $g_t(\mathcal{U}_t)$, and $Y_{\mathcal{U}_t}$ (Eqn.~\ref{eqn:hh}).
\end{algorithmic}
\end{algorithm}

\subsection{Detecting Hidden Heterogeneity}

Algorithm~\ref{alg:hh} provides a method to compute the 
\textit{detectable} hidden heterogeneity for a
region $\mathcal{U} \subseteq \mathcal{C}$ given a labeled calibration
data set $\mathcal{C}$ sampled from the same distribution as the test
set.  HH is calculated as the potential improvement (compared to the 
original classifier) achieved by training a specialized classifier on
only the items in $\mathcal{U}$.

In step 1, we define item $x_t$'s probability neighborhood
$\mathcal{U}_t$ to contain calibration items that are close to $x_t$ in
the probability simplex $\Delta_{K-1}$.  More precisely,
$\mathcal{U}_t$ contains the items within radius $r$ of item $x_t$ in
$\Delta_{K-1}$.
There is no {\em a priori} best choice for $r$, but to obtain
reliable HH estimates, one should choose $r$ such that no set
$\mathcal{U}_t$ is excessively small.
We employ the standard choice of Hellinger distance $D_H$
(Equation~\ref{eqn:hellinger}) to calculate the distance between 
probability vectors $\hat{p}_i$ and $\hat{p}_j$.  Hellinger distance
is the probabilistic equivalent of Euclidean distance,
and it is more suitable here than KL divergence, which is
not symmetric.
\begin{equation} \label{eqn:hellinger}
  D_H(\hat{p}_i, \hat{p}_j) = \frac{1}{\sqrt{2}} \sqrt{\sum_{k=1}^K
    \left(\sqrt{\hat{p}_i[k]} - \sqrt{\hat{p}_j[k]}\right)^2}.
\end{equation}
  
Conveniently, the Hellinger distance can be expressed as the Euclidean norm of the
difference of the element-wise square root of each probability
vector~\citep{krstovski:hellinger13}:
\begin{equation}
  \label{eqn:hell}
  D_H(\hat{p}_i, \hat{p}_j) = \frac{1}{\sqrt{2}} \left\lVert
  \sqrt{\hat{p}_i} - \sqrt{\hat{p}_j} \right\rVert_2.
\end{equation}
This in turn allows the use of efficient methods (e.g., k-d tree) for
populating neighborhood $\mathcal{U}_t$.

For each test item $x_t$, a new (local) classifier $g_t$ is trained
using only the nearby calibration items in $\mathcal{U}_t$ (step 2).
This classifier $g_t$ can be any classifier type.  We employed an
ensemble method that can perform internal generalization estimates
without an additional validation set.  We trained a bagged ensemble of
50 decision trees with no depth limit and no limit on the number of features
searched for each split.  We used out-of-bag error to determine
how much pruning to employ to achieve good generalization and avoid
overfitting to the calibration set.  We searched over 7 values of the
$\alpha$ pruning complexity parameter, evenly spaced between 0.0 (no
pruning) and 0.03, as input to the minimal cost-complexity pruning
method~\citep{breiman:cart84}.%

Finally (step 6), we calculate HH for $\mathcal{U}_t$ by 
comparing the Brier score~\citep{brier:score50} of the original
predictions by model $f$ on $\mathcal{U}_t$ (step 3) with those
generated by the local model $g_t$ (step 4) using true labels
$Y_{\mathcal{U}_t}$ (step 5): 
\begin{equation}
  \label{eqn:hh}
  \HH_{\mathcal{U}_t} = \text{Brier}(f(\mathcal{U}_t), Y_{\mathcal{U}_t}) -
  \text{Brier}(g_t(\mathcal{U}_t), Y_{\mathcal{U}_t}), 
\end{equation}
where the Brier score is the mean squared error between predictions
$\hat{p}_i[k] \in [0,1]$ and labels $y_i$, for $N$ items and $K$
possible classes: 
\begin{equation}
  \label{eqn:brier}
  \text{Brier}(\hat{p}, Y) = \frac{1}{N} \sum_{i=1}^{N} \sum_{k=1}^{K}
  \Big( \hat{p}_i[k] - \mathbb{1}(y_i = k) \Big)^2.
\end{equation}

We enforce the condition that $g_t$ is no worse than $f$ by clipping
$\HH_{\mathcal{U}_t}$ to 0.  Regions with large HH values provide
both a warning that global calibration methods may not perform well
and an opportunity for local specialization by using item similarity
during calibration.

\section{Similarity-Weighted Calibration}
\label{sec:swc}

We propose to improve point-wise calibration by leveraging 
information in feature space as well as the uncalibrated probabilities
$\hat{p}_i$.  Given test item $x_t$, the goal is to estimate
well-calibrated $\hat{q}_t[k] = P(y = k | x_t)$ for each class $k \in
\{1 \ldots K\}$.  Similarity-Weighted
Calibration (SWC) is described in Algorithm~\ref{alg:swc}.  Let
\[
s(t,i)=\text{sim}([x_t, \hat{p}_t], [x_i, \hat{p}_i]) \in [0,1]
\]
be the similarity between item $x_t$ and item $x_i$ measured in the
augmented space $\mathcal{X}^+$, where $[a, b]$
  is the concatenation of vectors $a$ and $b$. A similarity of 1 is perfect
identity.
The investigator chooses how best to measure similarity in this
space.  Use of $\mathcal{X}^+$ enables calibration to benefit from
information encoded by the classifier ($\hat{p}$) as well as item
position in feature space ($x$).  
Further, a supervised similarity measure can learn the relative
importance of each component for the problem at hand.
We employ such a measure: the random forest {\em proximity function} (RFprox).
RFprox trains a random forest on a labeled data set and defines the
similarity between items $x_i$ and $x_j$ as the fraction of times they
are assigned to the same leaf in each tree of the
ensemble~\citep{breiman:rf01,cutler:rf12}.  
Effectively, the random forest encodes
a ``kernel'' defined by those weights (leaf co-occurrences)~\citep{htf:ml09}. 
We employ the calibration data to learn the relevant RFprox
measure using a random forest with 100 trees, no depth limit, and
considering a random set of $\sqrt{d}$ features for each split, given
$d$ total features. 
Note that defining sim() using standard kernels, such as the Gaussian
kernel, over $\mathcal{X}^+$  would impose a fixed weighting on the
$x$ and $\hat{p}$ components. A potential direction for future
research would be to apply multiple kernel learning \citep{Gonen2011}
to optimally combine separate kernels for  $x$ and $\hat{p}$.

SWC computes the similarity of $x_t$ to every item in the calibration
set (step 3) and uses this information to replace $\hat{p}$ with a
similarity-weighted combination of labels from the calibration set
(step 4). 
\begin{equation}
  \label{eqn:simsum}
  \hat{q}_t[k] = \frac{1}{\sum_i s(t, i)}
  \sum_i s(t, i) \mathbb{1}(y_i = k).
\end{equation}

\begin{algorithm}[tb]
\caption{Similarity-Weighted Calibration (SWC)}
\label{alg:swc}
\textbf{Input}: Test item $x_t$, calibration data $x_i \in \mathcal{C}$ and
labels $y_i$ \\
\textbf{Output}: Calibrated probabilities $\hat{q}_t[k], \forall k$
\begin{algorithmic}[1] 
  \STATE Collect model predictions for item $x_t$: $\hat{p}_t[k]$ for
  $k \in \{1 \ldots K\}$.
  \STATE Collect model predictions for the calibration set: $\hat{p}_i[k]$ for
  $x_i \in \mathcal{C}, k \in \{1,\ldots,K\}$.
  \STATE Compute pairwise similarity as $s(t, i)$ for $x_i \in
  \mathcal{C}$. 
  \STATE Compute $\hat{q}_t[k] = \frac{1}{\sum_i s(t, i)}
    \sum_i s(t, i) \mathbb{1}(y_i = k)$ for $k \in
  \{1,\ldots,K\}$ (Eqn.~\ref{eqn:simsum}).
\end{algorithmic}
\end{algorithm}

\comment{
As noted above, SWC differs from SBA-10 in that it uses a weighted sum
across all items in the calibration set rather than averaging the
labels of only the 10 nearest
neighbors.  Like SBA-10, similarity is computed in augmented feature
space.  The choice of similarity measure is a key ingredient in any
similarity-based calibration method.  SBA-10 uses the Weka IBk
implementation of the k-nearest-neighbors algorithm,
which by default employs Euclidean distance.
}

The similarity-based approach to calibration enables local
specialization within the data set, but it does not directly make use
of the calculated HH.  We also developed the SWC-HH algorithm, which
filters the calibration set to restrict which items are used to
generate $\hat{q}$.  HH, which is computed separately for each test
item $x_t$ (Algorithm~\ref{alg:hh}), is
employed as an additional filter for calibration.   
In step 4, SWC-HH only includes items with a similarity
to item $x_t$ of at least $\frac{1}{2}{\HH_{\mathcal{U}_t}}$.  Note
that the maximum value for HH is 2.0 since it is the
difference in Brier scores, clipped to 0.0, and each Brier score
ranges between $0.0$ and $2.0$.  Dividing by 2.0 normalizes the
threshold to the range $0.0$--$1.0$, making it suitable as a similarity
threshold.

\comment{Global methods assume
that all items with the same predicted probability require the same
correction, while local methods allow the possibility that one
prediction is over-confident while another is under-confident, even if
they have the same predicted probability.}

\section{Experimental Results}
\label{sec:res}

We conducted experiments with a variety of classifiers and data sets
to compare local and global calibration methods and to determine the
role that hidden heterogeneity plays.
Our hypotheses were that (1) local, similarity-based calibration is
more effective at reducing Brier score than global calibration
methods, (2) the amount of improvement correlates with the
hidden heterogeneity score, and (3) calibration support can serve as an
indicator of per-item calibration quality.  Our implementations of SWC,
SWC-HH, and other 
calibration methods, along with scripts to replicate the experiments,
are available at \url{https://github.com/wkiri/simcalib}.

\subsection{Methodology}

We assessed calibration methods for six tabular data classifiers
as implemented in the scikit-learn Python library~\citep{scikit-learn},
including
a decision tree with {\tt min\_samples\_leaf = 10} (DT), %
a random forest with 200 trees (RF), %
an ensemble of 200 gradient-boosted trees (GBT), %
a linear support vector machine (SVM), %
a Gaussian kernel ($\gamma=\frac{1}{d \: var(X)}, C=1.0$) support
vector machine (RBFSVM), %
and a Naive Bayes classifier (NB).  Any parameters not explicitly
mentioned were set to their default values.
We also conducted experiments with
pre-trained deep neural networks for classifying images (details
in Section~\ref{sec:image}).

\paragraph{Data sets.}  We analyzed four tabular and two image data sets:
\begin{itemize}
  \vspace{-0.15in}
  \item moons: a synthetic 2D data set with two partially
    overlapping classes, chosen to enable visualization of classifier
    outputs and hidden heterogeneity in feature space. 
    Observations were generated using the scikit-learn {\tt make\_moons()}
    function with {\tt noise} set to 0.3 and {\tt random\_state} set
    to 0.
  \vspace{-0.1in}
  \item letter (letter recognition): a 26-class data set of capital
    letters from the 
    English alphabet represented by 16 statistical and geometrical
    features that describe the image of the
    letter~\citep{frey:letter91},
    available at
    \url{https://archive.ics.uci.edu/ml/datasets/letter+recognition}.
  \vspace{-0.1in}
  \item mnist: the MNIST handwritten digit data set~\citep{lecun:mnist98}
    composed of 28x28 pixel ($d=784$) images containing a digit from 0 to
    9.  We used the data set as provided by OpenML; the original
    source is \url{http://yann.lecun.com/exdb/mnist/}.  In addition to
    the 10-class data set (mnist10), we created several binary subsets
    consisting only of two digits, such as ``4'' and ``9''
    (mnist-4v9).  This data set is ``tabular'' (1-d feature vector); 
    no 2D image structure is preserved.
  \vspace{-0.1in}
  \item fashion-mnist: grayscale images of
    clothing and accessories (10 classes) that was
    designed to be more challenging than the MNIST data set yet have
    the same dimensionality (28x28, $d=784$)~\citep{xiao:fmnist17},
    available at
    \url{https://github.com/zalandoresearch/fashion-mnist}. 
  \vspace{-0.1in}
\item CIFAR-10: \num{60000} images (64 $\times$ 64 pixels) labeled into
  10 distinct classes; the test set contains \num{10000} images~\citep{cifar10}
  \vspace{-0.1in}
\item CIFAR-100: a disjoint set of \num{60000} images labeled into
  \num{100} different classes ($50$k train, $10$k test)
  \vspace{-0.1in}
\end{itemize}

For tabular data sets, we randomly sampled \num{10000} items
and for each trial randomly split them into \num{500} train, \num{500}
test, and 
\num{9000} for a calibration pool.  For the ``mnist10'' and ``letter''
data sets, we used \num{1000} items each for training and test, due to
their large number of classes (10 and 26, respectively).  We
created a series of nested calibration sets
of size \{ 50, 100, 200, 500, 1000, 1500, 2000, 2500, 3000 \} to
assess the data efficiency of each calibration method. 
For image data sets, in each trial we generated a class-stratified
random split of the standard test set into \num{5000}
test items and reserved the remainder as the calibration set.
We report average performance across 10 trials along
with the standard error for the observed values.

\paragraph{Calibration methods.}  We compared three similarity-based
calibration methods (SBA-10, SWC, and SWC-HH) to standard calibration methods
including Platt scaling~\citep{platt:calib99}, histogram
binning~\citep{zadrozny:hist01}, and isotonic
regression~\citep{zadrozny:iso02}.  
%
SBA-10 employed
Euclidean distance to identify the nearest neighbors, while
SWC and SWC-HH used the RFprox similarity measure.
When computing hidden heterogeneity, we used a probability
radius of $r=0.1$ in the probability simplex. 

Platt scaling optimizes the negative
log-likelihood of predictions against the target probabilities rather
than discrete $\{0,1\}$
labels~\citep{platt:calib99,niculescu-mizil:calib05}.  With $n_{+}$ as
the number of calibration items in the positive class and $n_{-}$ as
the number of negative items, the target probabilities are
$y_{+} = \frac{n_{+} + 1}{n_{+} + 2}$ 
and
$y_{-} = \frac{1}{n_{-} + 2}$ .
For multi-class problems, we applied temperature
scaling~\citep{guo:calib17}.
For classifiers that output
probabilities instead of logits, we first transformed $\hat{p}_i$ into
logits as $z_i[k] = \ln \frac{\hat{p}_i[k]}{1 - \hat{p}_i[k]}$.  To avoid
dividing by zero (or taking its logarithm), we clipped
$\hat{p}_i[k]$ to the range $[\epsilon, 1 - \epsilon]$, where $\epsilon =
1 \times 10^{-12}$.
We applied the histogram binning method implemented
by~\citet{kumar:calib19}\footnote{Available at
\url{https://github.com/p-lambda/verified_calibration}.} and followed
their recommendation to use equal-mass bins (100 bins).

\comment{
Finally, we assessed a hybrid method that first applies global
calibration to address global over- or under-confidence and then
refines the probabilities by applying BSC.  We refer to this method as
Platt+BSC or TS+BSC to indicate the use of Platt or temperature
scaling, depending on the number of classes.
}

\paragraph{Metrics.}  We employ Brier score (Equation~\ref{eqn:brier})
to measure point-wise prediction quality, following earlier
researchers~\citep[e.g.,][]{zadrozny:hist01,zadrozny:iso02,niculescu-mizil:calib05}.
It has several advantages over a commonly used calibration metric
called the Expected Calibration Error 
(ECE)~\citep{naeini:ece15}, which assigns
predictions to bins to assess aggregate calibration.  ECE can be trivially
minimized to 0 by always predicting the empirical average probability of a
given class, yielding perfectly calibrated but uninformative
predictions~\citep{kull:beta-calib17,widmann:calib19,ovadia:calib19}.
The Brier score incorporates not just
calibration error (or ``reliability'') but also ``resolution'' or
sharpness, which rewards predictions that do more than predict the
average probability~\citep{ferro:debias12}.
In addition, ECE is sensitive to the number and choice of
bins~\citep{vaicenavicius:calib19,kumar:calib19,patel:mi-calib21},
it exhibits undesirable edge effects
(discontinuities at bin boundaries), and it only assesses calibration
of the predicted class. 
Brier score characterizes prediction quality across all classes, and
it avoids artificial discretization and edge effects, since it does not
employ binning.
\comment{
To likewise avoid the drawbacks of binning-based depictions
for reliability diagrams, including coarse granularity and
discontinuities at bin boundaries, we use the kernel density
estimation approach developed by~\cite{zhang:calib20}, which estimates
the accuracy for each test item using a triweight kernel in
probability space.
}

\begin{figure}
  \begin{center}
  \subfloat[Naive Bayes]{\includegraphics[width=1.9in]
    {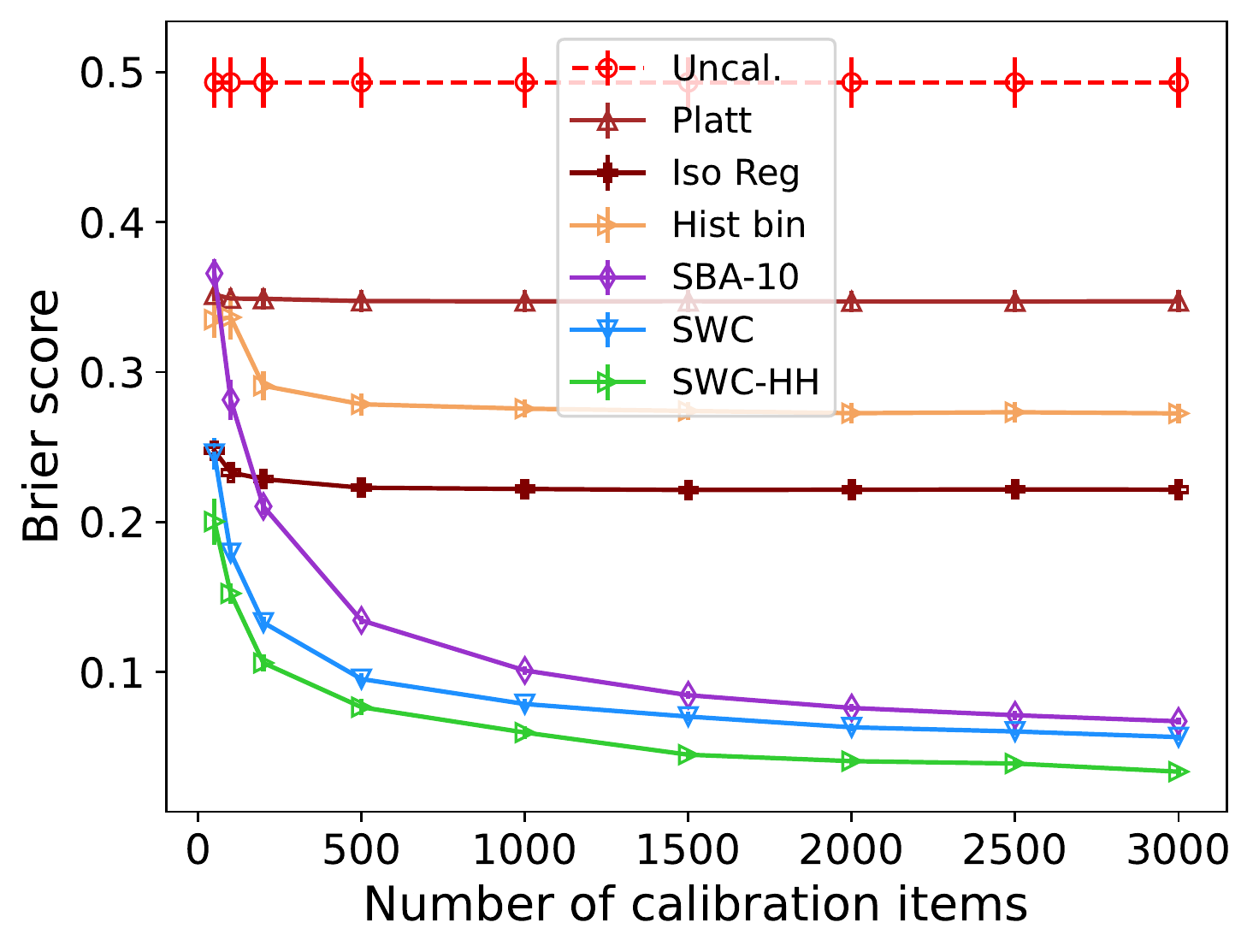}}
  \subfloat[Decision tree]{\includegraphics[width=1.9in]
    {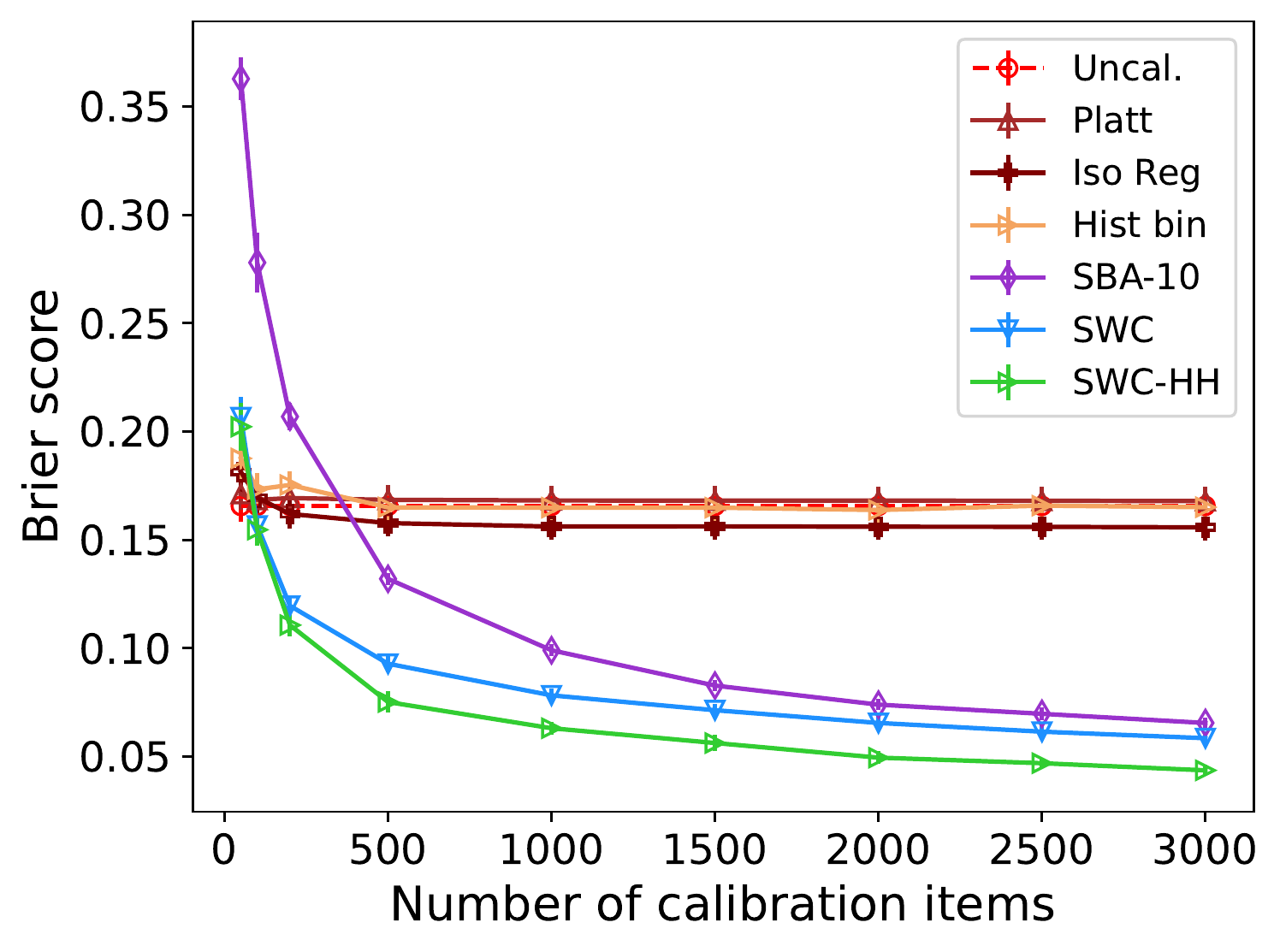}}
  \subfloat[Random forest]{\includegraphics[width=1.9in]
    {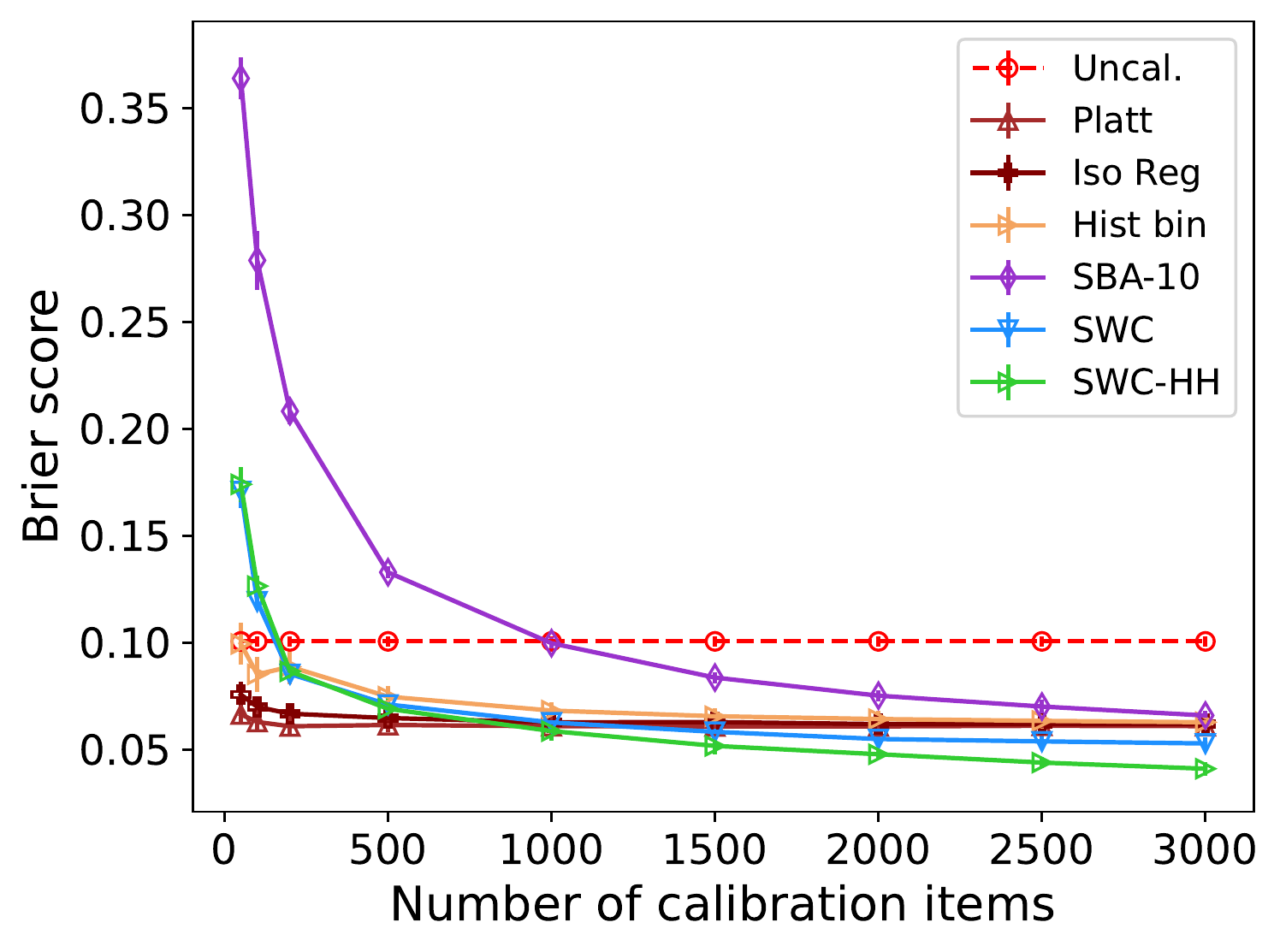}}
  \end{center}
  \caption{Calibration performance for three
    classifiers on the MNIST 4-vs-9 data set.  Each plot shows Brier
    score (lower is better) with increasing calibration set size.
    Error bars show one standard error over 10 trials.}
  \label{fig:varycal}
\end{figure}

\subsection{Similarity-based calibration for tabular data sets}
\label{sec:loc}
 
To test our first hypothesis, we
compared similarity-based calibration to global methods. 
Figure~\ref{fig:varycal} shows Brier score as a
function of available calibration data for the binary classification
task of distinguishing two handwritten MNIST digits (``4'' vs.~``9''). 
%
The uncalibrated predictions yielded different starting Brier scores
for each classifier (red dashed lines).
Platt scaling and isotonic regression improved the Brier score
for the Naive Bayes (NB) and 
random forest (RF) classifiers but only marginally for the decision
tree (DT).  No further improvements were achieved beyond 500
calibration items.
Similarity-based calibration (SBA-10, SWC, and SWC-HH)
generally did not perform well with small calibration sets but
achieved much
larger benefits for NB and DT when at least 500 items were used for
calibration, and Brier score continued to improve as more calibration
data was provided.  The random forest, which had the best initial Brier
score and therefore less room for improvement, showed an advantage for
similarity-based calibration after at least 1500 items were used.
SWC-HH yielded a clear advantage at all calibration set
sizes for NB and DT, and it also provided a small advantage over SWC
for RF with calibration set sizes of at least 1500 items.

SWC and SWC-HH consistently out-performed SBA-10.
Since RFprox uses labels to
learn the similarity measure, it can elevate the importance of
individual features in $\mathcal{X}^+$ (like $\hat{p}[k]$) by placing
them higher in individual decision trees within its ensemble.  
SWC and SWC-HH also include evidence from the entire calibration set
rather than only the nearest neighbors.
Importantly, SBA-10 showed almost no
difference in Brier score for different classifiers, effectively
ignoring their individual differences in $\hat{p}$.  MNIST is
represented by 784 features, so the addition of two dimensions in
$\mathcal{X}^+$ has little
effect.  In contrast, the fact that SWC obtained different absolute
results for each classifier indicates that RFprox
effectively leveraged the $\hat{p}$ features.
Experimental results for all classifiers and all tabular data sets,
reporting Brier score and accuracy results, are provided in
Appendix~\ref{app:res} (Figures~\ref{fig:res-tabular-binary}
and~\ref{fig:res-tabular-multi}).  
  
\begin{figure}
  \begin{center}
  \subfloat[Platt scaling (4 binary)]{\includegraphics[width=2in]
    {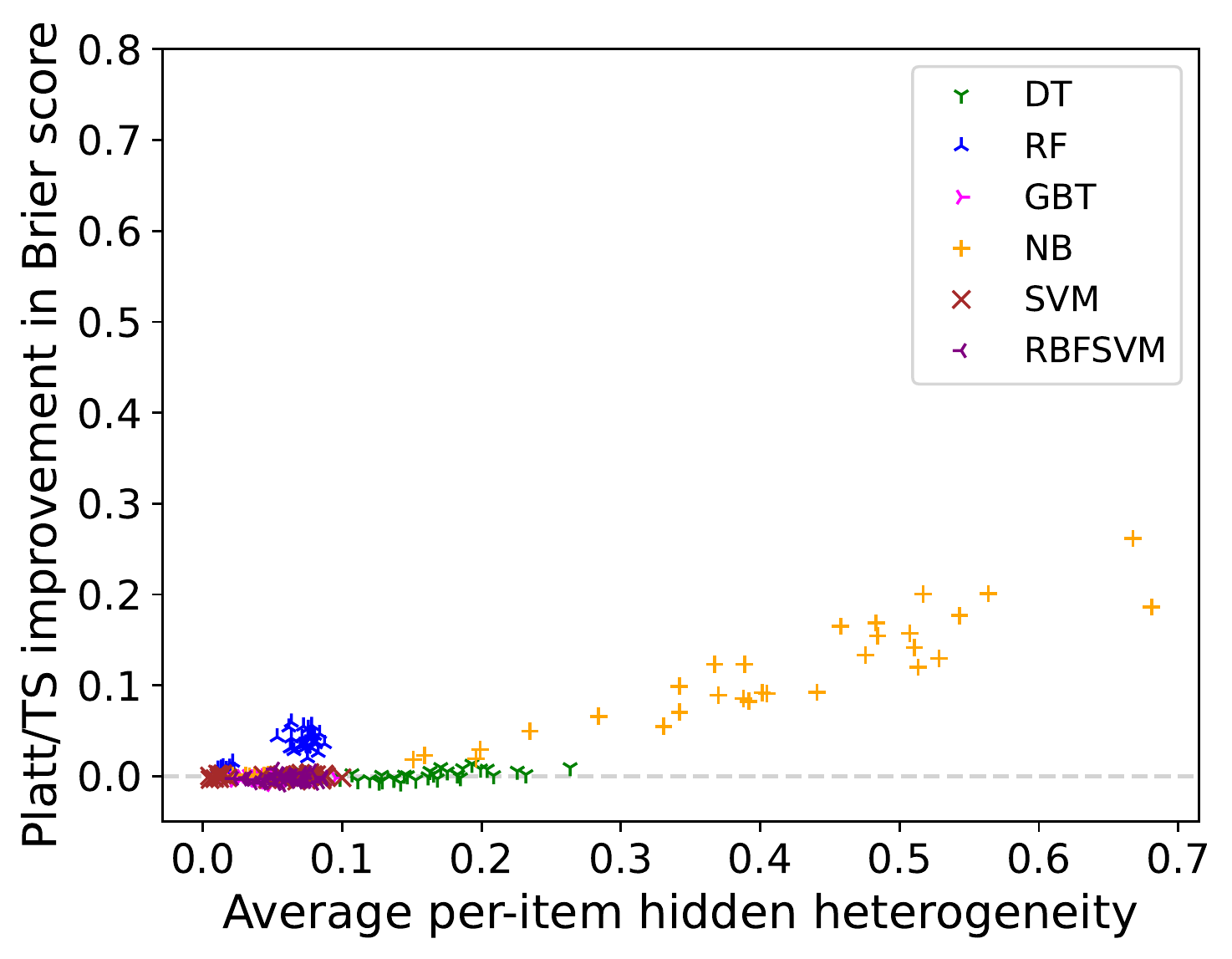}}
  \subfloat[SWC (4 binary)]{\includegraphics[width=2in]
    {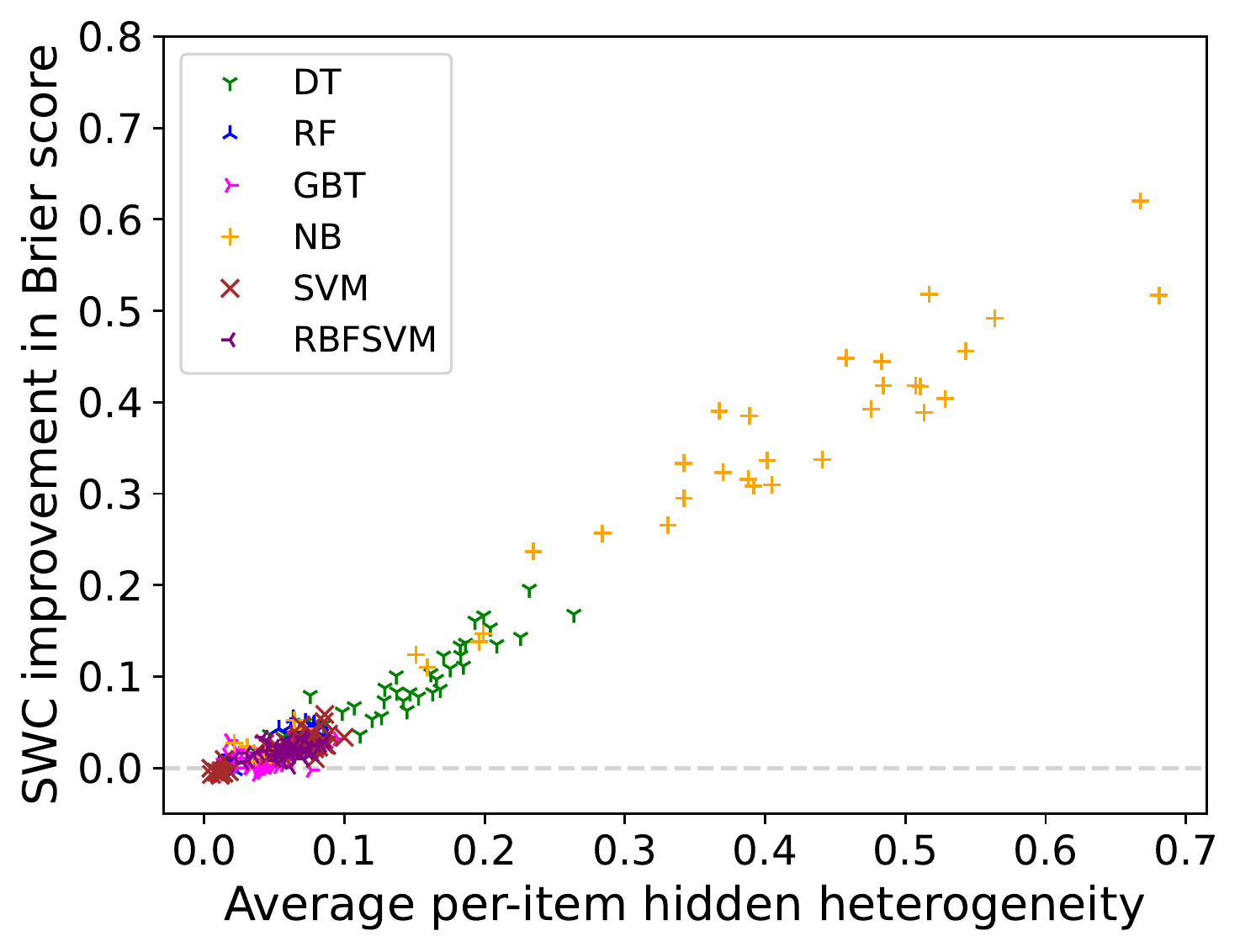}}
  \subfloat[SWC-HH (4 binary data sets)]{\includegraphics[width=2in]
    {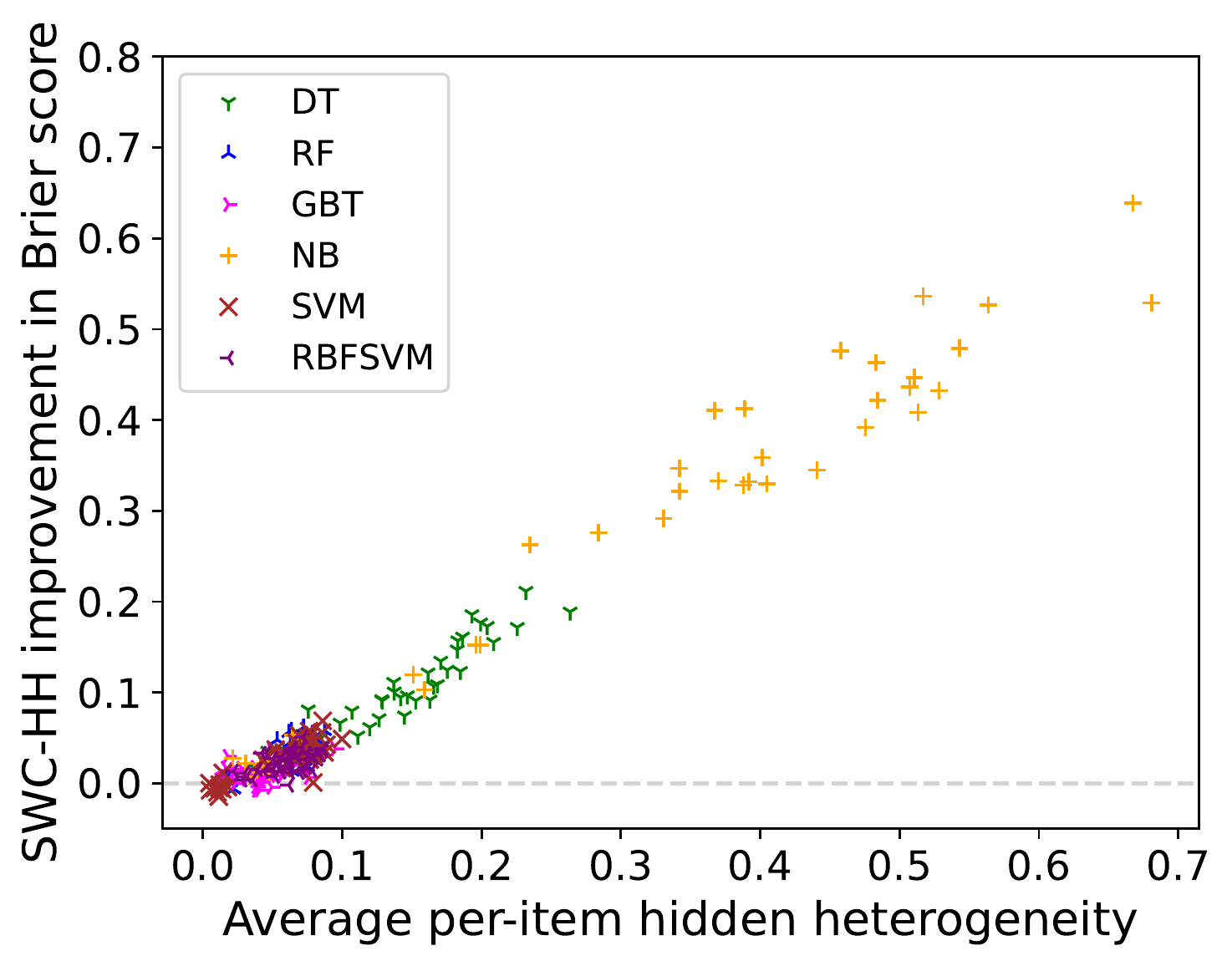}}
  \\
  \subfloat[Temperature scaling (3 multi-class)]{\includegraphics[width=2in]
    {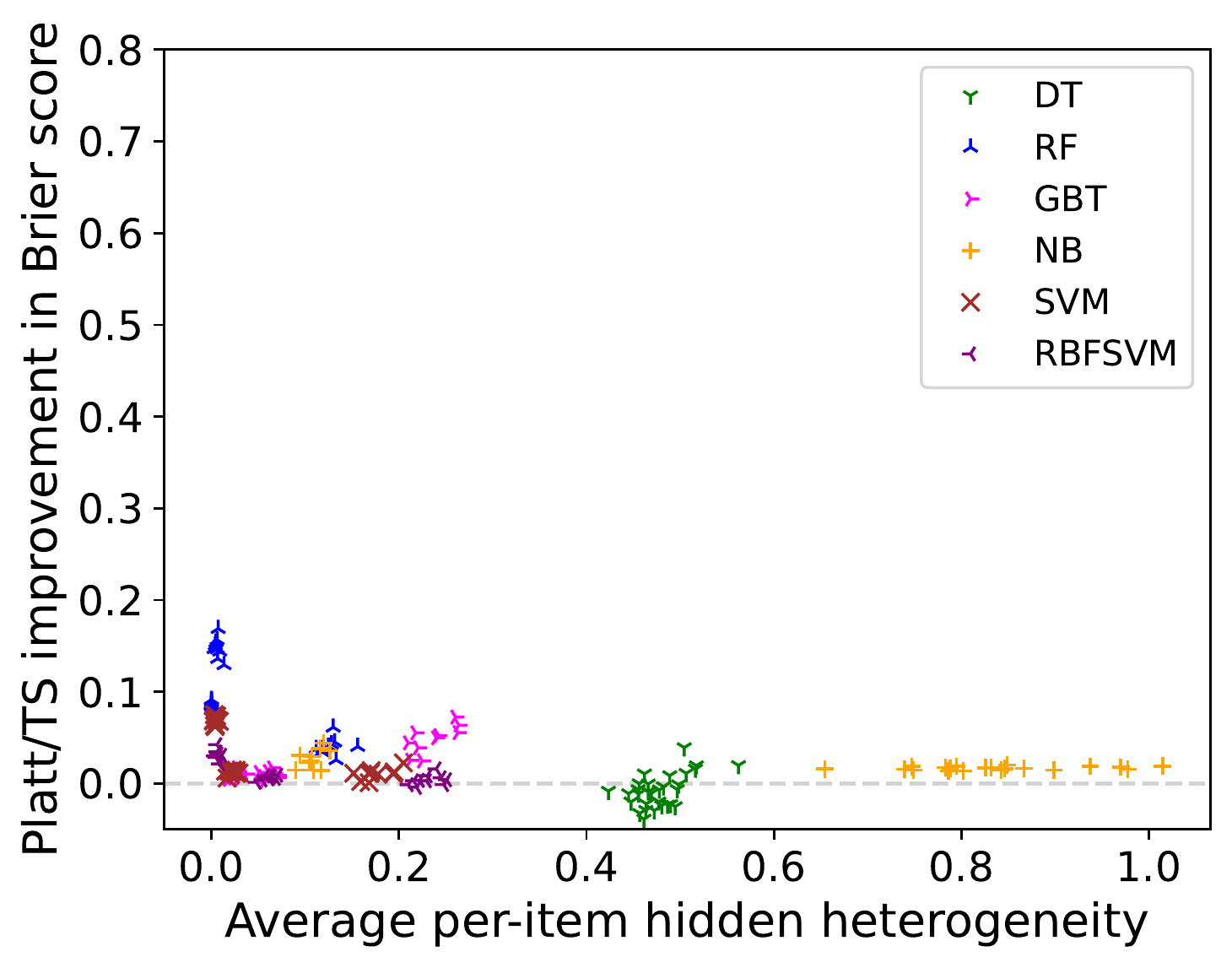}}
  \subfloat[SWC (3 multi-class)]{\includegraphics[width=2in]
    {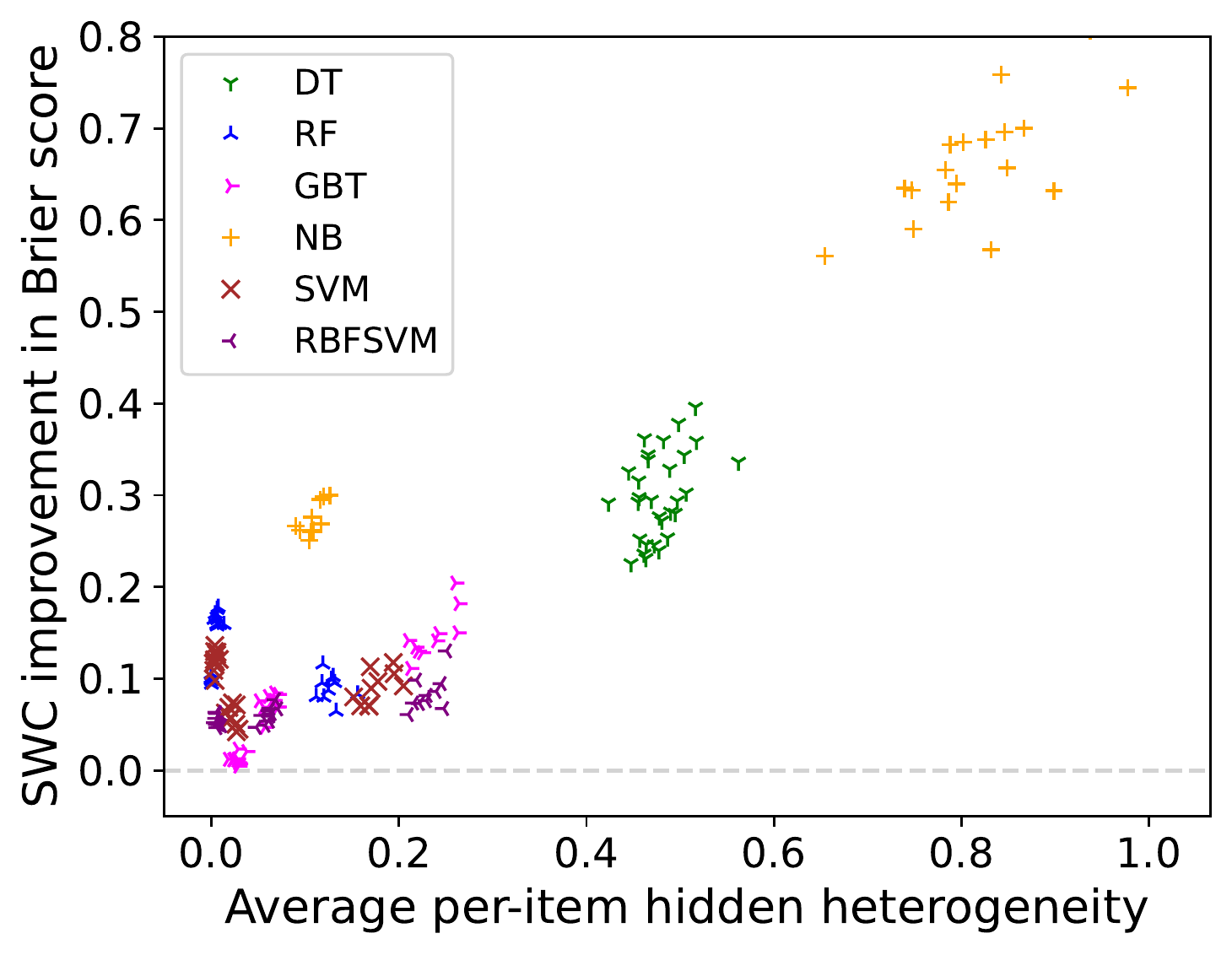}}
  \subfloat[SWC-HH (3 multi-class data sets)]{\includegraphics[width=2in]
    {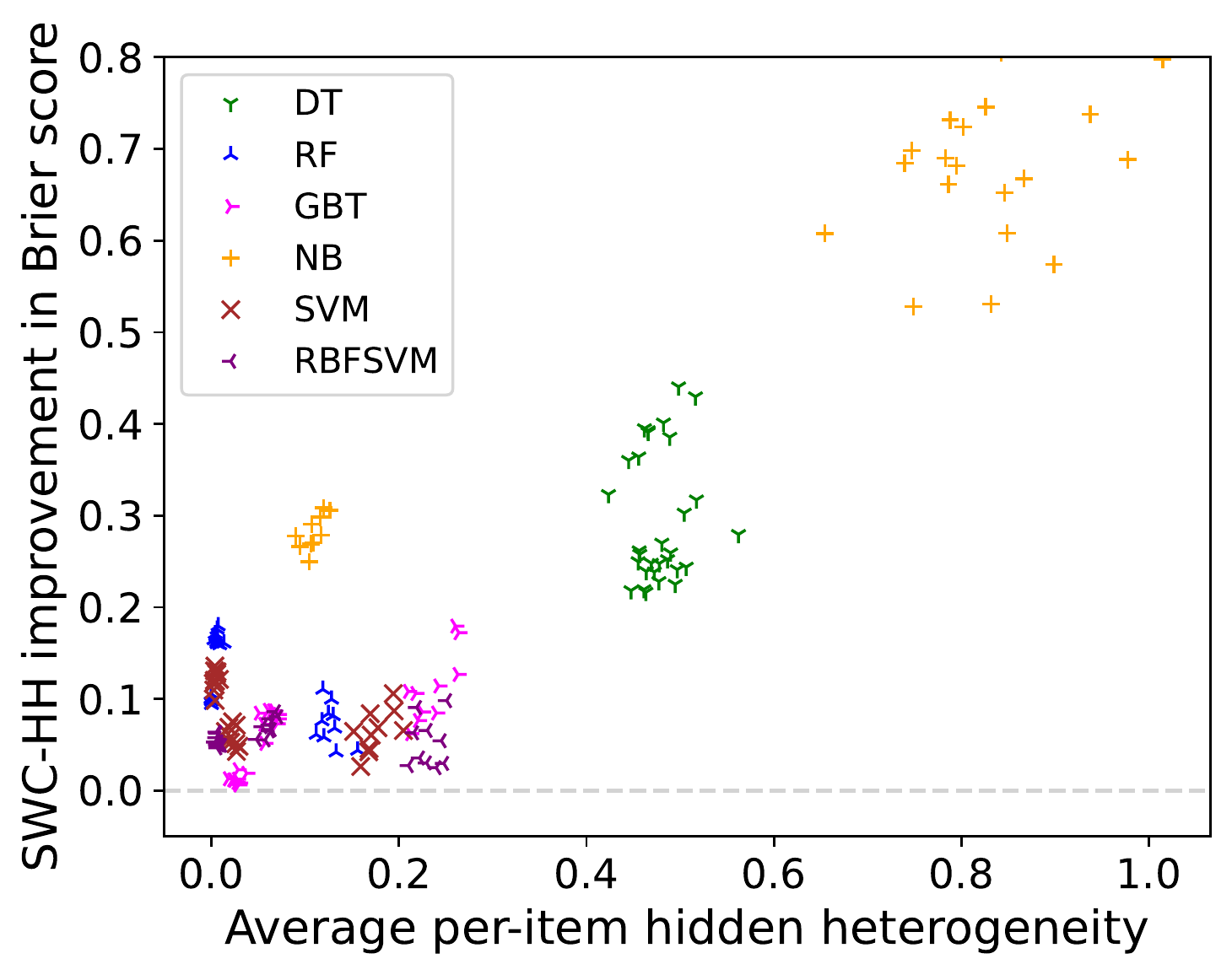}}
  \end{center}
  \caption{Brier score improvement versus average hidden
    heterogeneity for four binary (top row) and three multi-class
    (bottom row) tabular data sets,
    for six classifier types 
    and 10 random trials.}
  \label{fig:bs_vs_hh}
\end{figure}

\subsection{Similarity-based calibration exploits hidden heterogeneity}

Our second hypothesis was that HH helps explain why and when
similarity-based calibration is effective.  We examined Brier score
improvement across a large combination of data sets, classifiers, and
random trials.
We found that large HH can be present even in the absence of domain
shift, which creates a large opportunity for local calibration.
Figure~\ref{fig:bs_vs_hh} shows the improvement (reduction) in Brier 
score after calibration as a function of the average HH across the
test set. 
The four binary data sets were MNIST ``1'' vs. ``7'' (relatively
easy), ``4'' vs.~``9'' and ``3'' vs.~``8'' (intermediate), and ``3''
vs.~``5'' (difficult).
The three multi-class data sets were ``mnist10'', ``fashion-mnist'',
and ``letter''. 
We compared Platt (for binary) or
temperature scaling (multi-class) calibration to SWC and SWC-HH
for six classifiers, 
using 10 trials per combination of data set and classifier.
SWC and SWC-HH achieved Brier score improvements that correlate strongly 
with the amount of measured HH. 
The relationship was far weaker for the global Platt/temperature
scaling methods, which cannot detect or exploit HH.
%
%
These results show that HH, which can be computed prior to
calibration, is a useful diagnostic indicator that can guide 
the choice of calibration strategy.  When HH
is high, it is advisable to employ a similarity-based calibration
method like SWC.  When HH is low or there is not much calibration data, global
methods such as Platt scaling are recommended.

\begin{figure}
  \centering
  \includegraphics[width=6in]{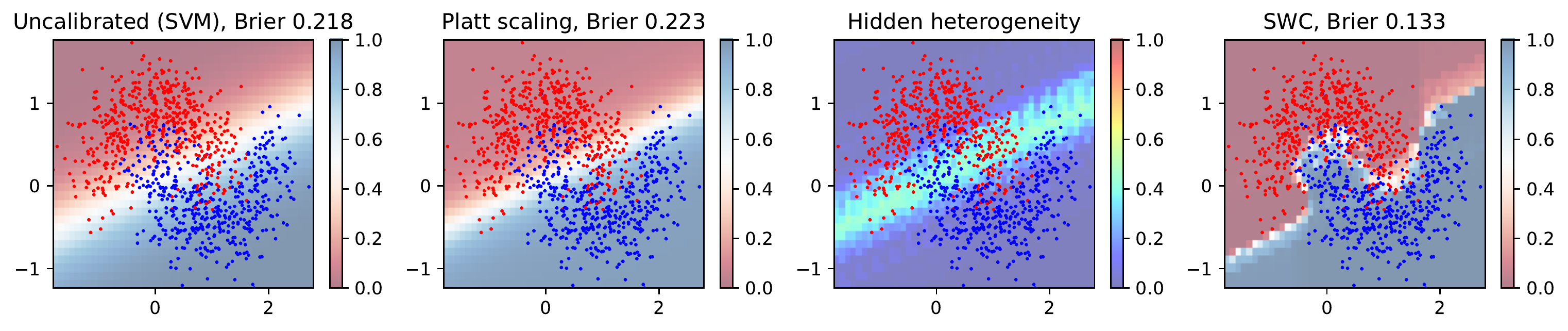}
  \includegraphics[width=6in]{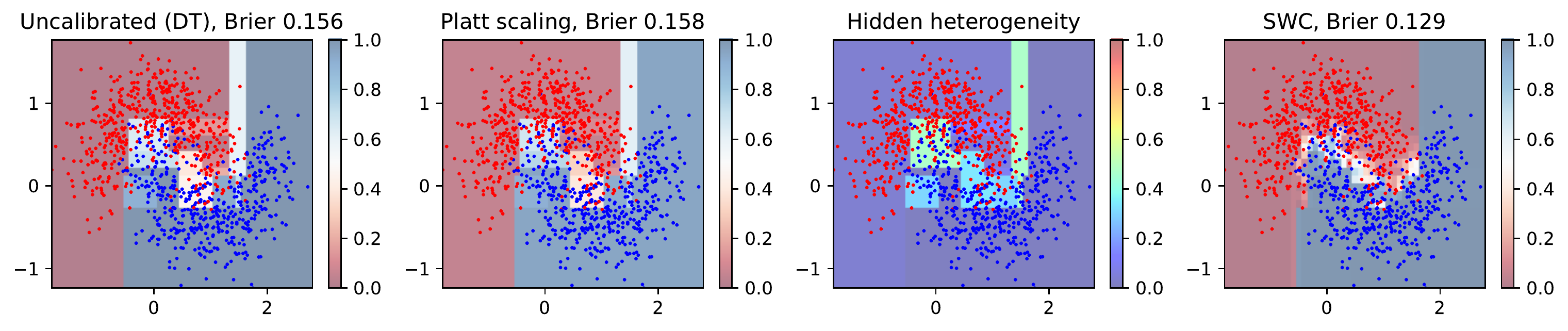}
  \includegraphics[width=6in]{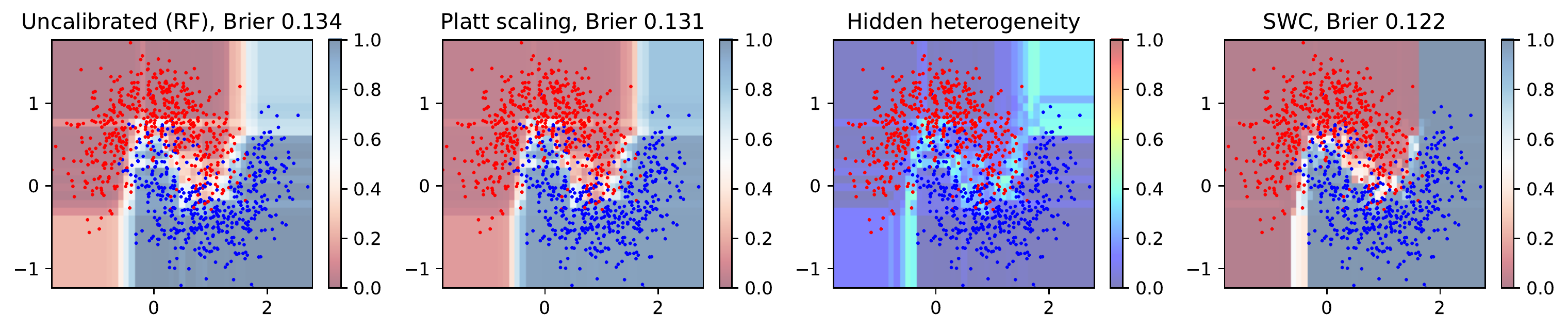}
  \caption{Visualization of uncalibrated predictions (first column)
    for the ``moons'' data set, Platt scaling (second column), and
    SWC (fourth column).  The third column shows hidden
    heterogeneity values that highlight areas of potential
    improvement, which align with SWC improvements.}
  \label{fig:viz-HH}
\end{figure}

To better understand how SWC and SWC-HH improve Brier score,
we visualize the calibration process in Figure~\ref{fig:viz-HH} for
the two-dimensional ``moons'' data set.  Each row corresponds to
results for a different classifier (linear SVM, decision tree, and
random forest).  The third column shows the computed HH values.
The classifiers were trained on 500 points, calibrated using 1000
points, and Brier score evaluated on 500 points.
The linear SVM cannot model the
nonlinear decision boundary very well.  The diagonal region where the classes
are mixed yet separable in the feature space has a high value for HH.
When SWC is applied, it is the $\hat{p}$ values in this band that
receive the biggest modifications.  These changes reduce (improve) the
Brier score from \num{0.218} to \num{0.133}.  
Platt scaling increases (worsens) the Brier score slightly. 
The decision tree (second row of Figure~\ref{fig:viz-HH}) exhibits
less hidden heterogeneity on the same data set, because it is able to
model the nonlinear decision boundary more effectively.  SWC improves
the Brier score from \num{0.156} to \num{0.129}.
Finally, the random forest (bottom row of Figure~\ref{fig:viz-HH})
exhibits more areas with hidden
heterogeneity, due to overly conservative predictions in the
  upper right and lower left areas.  SWC creates smoother regions as the
calibration data informs updates to the posterior probabilities
and improves the Brier score from \num{0.134} to \num{0.122}.  

Importantly, the difference in results for the rightmost column in
Figure~\ref{fig:viz-HH} demonstrates that SWC adapts (calibrates) most
where the classifier exhibits hidden heterogeneity, yielding a result
that is customized to the original classifier and more flexible than
global calibration.

\subsection{Similarity-based calibration for image classifiers}
\label{sec:image}

We also conducted calibration experiments with the CIFAR-10 and CIFAR-100
data sets \citep{cifar10} using three pre-trained neural networks of
increasing
complexity\footnote{Pre-trained models were obtained from
\url{https://github.com/chenyaofo/pytorch-cifar-models} .}.
ResNet20 \citep{he:resnet16} has 20 layers and 0.27M parameters,
ResNet56~\citep{he:resnet16} has 56 layers and 0.85M parameters,
and RepVGG\_A2~\citep{ding:repvgg21} has 22 layers and 25.49M
parameters.
%
For these data sets, the similarity measure $sim$ used by SWC and
SWC-HH operates in the latent 
space learned by each network.  Specifically, we used the output
activations of the {\tt avgpool} 
(for ResNet models) and {\tt gap} (for RepVGG\_A2)
layers as a feature vector (dimensionality \num{64},
\num{64}, and \num{1408} respectively).
We again used a learned RF proximity function to compute similarity
in this space.
We found that a probability radius of \num{0.05} yielded reasonably sized
neighborhoods for computing HH.

\begin{figure}
  \centering
  \subfloat[CIFAR-10]{\includegraphics[width=3.25in]
    {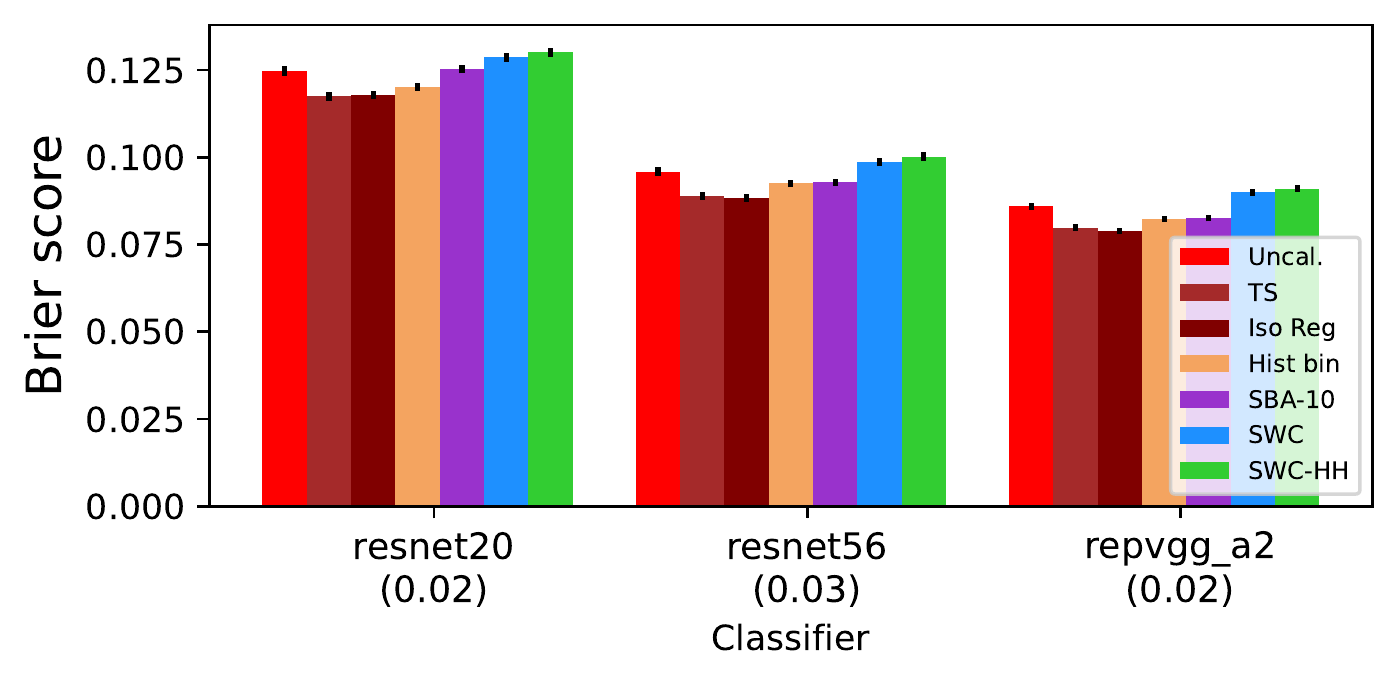}}
  \subfloat[CIFAR-100]{\includegraphics[width=3.25in]
    {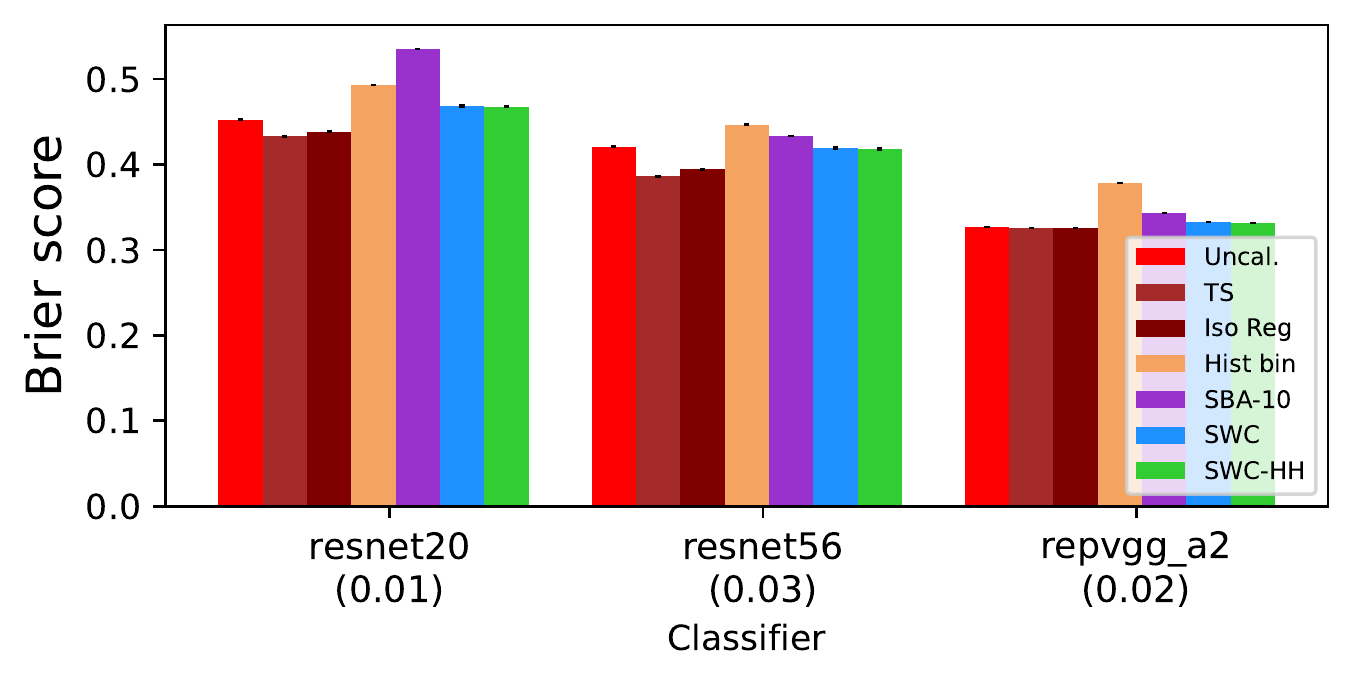}}
  \caption{Calibration performance on CIFAR-10 and CIFAR-100
    using three pre-trained neural networks over
    10 trials (error bars show standard error). HH is shown in parentheses.}
  \label{fig:res-cifar}
\end{figure}

Improvements for CIFAR-10 and CIFAR-100 are evident as more complex
neural networks are trained; the Brier score consistently decreases from
ResNet20 to 
ResNet56 to RepVGG\_A2 (see Figure~\ref{fig:res-cifar}).  However, HH
values were very low for all three networks ($0.01-0.03$), leaving
little room for improvement with local calibration.  Indeed, we found
that global calibration (temperature scaling or isotonic
  regression) yielded the best results 
for these data sets.  Consistent with the results on tabular data
shown in section~\ref{sec:loc}, calculating HH in advance provides
guidance as to which method will be most useful.

There is room for additional improvement.  Recent studies of deep
network latent spaces suggest that distances computed in neural
network latent spaces often do not work well. For example, nearest
neighbor classifiers using latent space distances perform
substantially worse than the standard multinomial logistic regression
(softmax) classifiers~\citep{garrepalli:oracle22}.  These latent spaces
are also not able to represent dimensions of variation that were
poorly sampled in the training
data~\citep{dietterich:familiarity22}. Likewise, decision tree
classifiers do not perform well on these learned
representations~\citep{garrepalli:oracle22}. Since HH is
  influenced by the data representation,
hidden heterogeneity could be more detectable for these data sets
using a different representation.  Likewise, even if HH is large,
similarity-based local 
calibration may not always be able to improve the Brier score due to
limitations of the representation.  
Employing t-SNE or PCA to reduce dimensionality for similarity
calculations (as done by~\cite{luo:local22}) could also be beneficial.
Exploring the connection between 
choice of representation and calibration efficacy is an important
future direction.


\comment{
Accuracy improvements for this data set range from 5--6\%, which is
equivalent to a year of state-of-the-art 
improvements on this data set.  The improved accuracy was achieved
effectively for free, since it did not require 
any additional data collection, model architecture updates, or
model innovations, while simultaneously reducing the Brier score.
}

\subsection{Calibration support highlights calibration data gaps and
  domain shift}

Our third hypothesis is that measuring calibration support, which is a
unique capability of similarity-based calibration methods, can provide useful
information about the relevance of the calibration set to each item
being calibrated.
%
We define the {\em calibration support} $S$ for item $t$ that
informs $\Phi(\hat{p} | t)$ as the sum of similarity weights
for items drawn from calibration set $\mathcal{C}$:
\begin{equation}
  S_{t, \mathcal{C}} = \sum_i s(t, i), x_i \in \mathcal{C}.
\end{equation}

Identifying items with low values for $S_{t, \mathcal{C}}$ can draw
attention to observations that are not well represented by the
calibration set.  These could be individual outliers or, if there is a
large number of such items, they could indicate distribution
shift between the calibration and test sets.  Low $S_{t, \mathcal{C}}$
values signal the need for more data (or more representative data) to
be added to $\mathcal{C}$.  While previous studies focus solely on
calibration performance as a function of the total calibration set
{\em size}, similarity-based calibration can characterize the
{\em relevance} of the calibration set to individual test items.

\begin{figure}
  \centering
  \includegraphics[width=6.5in]{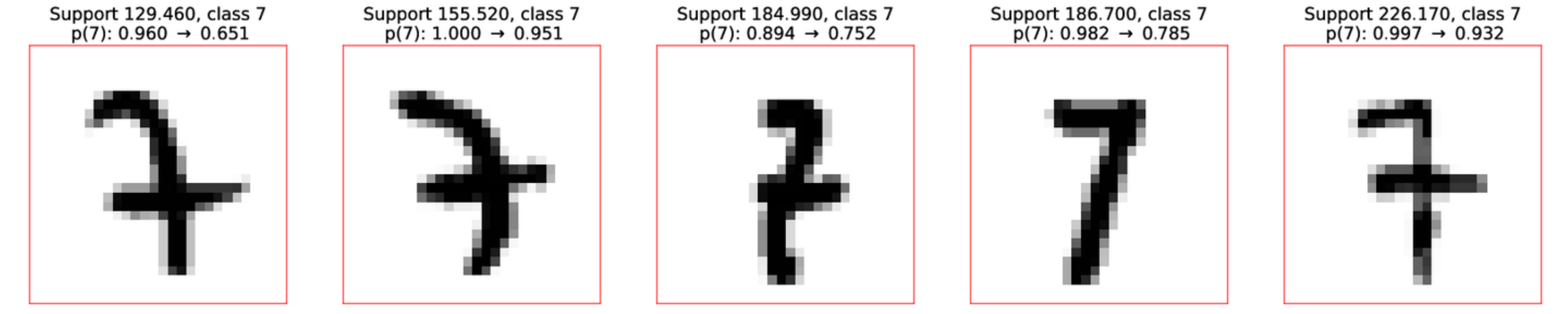}
  \caption{Test items from ``mnist-1v7'' with the lowest calibration
    support for a linear SVM classifier.  For all five items, SWC {\em
      reduced} the confidence of the correct class, instead of
    increasing it.}
  \label{fig:mnist-bot}
\end{figure}

Consider a linear SVM trained on \num{500} MNIST ``1'' and ``7''
digits and calibrated using SWC with \num{3000} digits.  For most
items in the test set, calibration improves.  However, examining
items with low calibration support helps us understand failures.
Figure~\ref{fig:mnist-bot} shows the five test items (out of \num{500}
total) with the lowest calibration support.
Calibration with SWC was detrimental
($\hat{q}[y]$ decreased) for all five.  These items are not
necessarily ambiguous in an objective sense, but the fact that they have
low representation in 
the calibration set signals (correctly) that the calibrated output
$\hat{q}$ may be less reliable.

\begin{figure}
  \centering
  \subfloat[0\% rotated]
           {\includegraphics[width=1.6in]{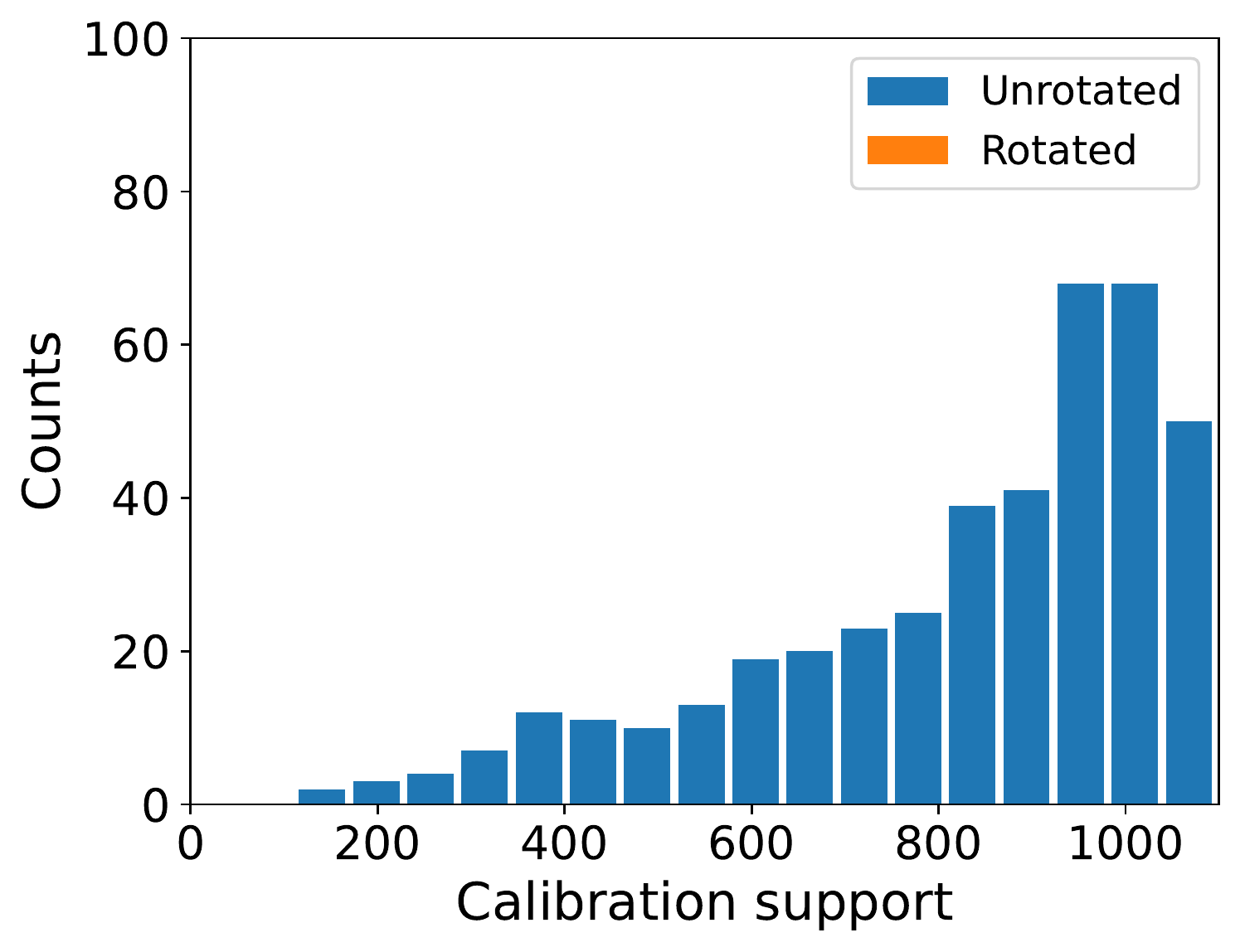}}
  \subfloat[10\% rotated]
           {\includegraphics[width=1.6in]{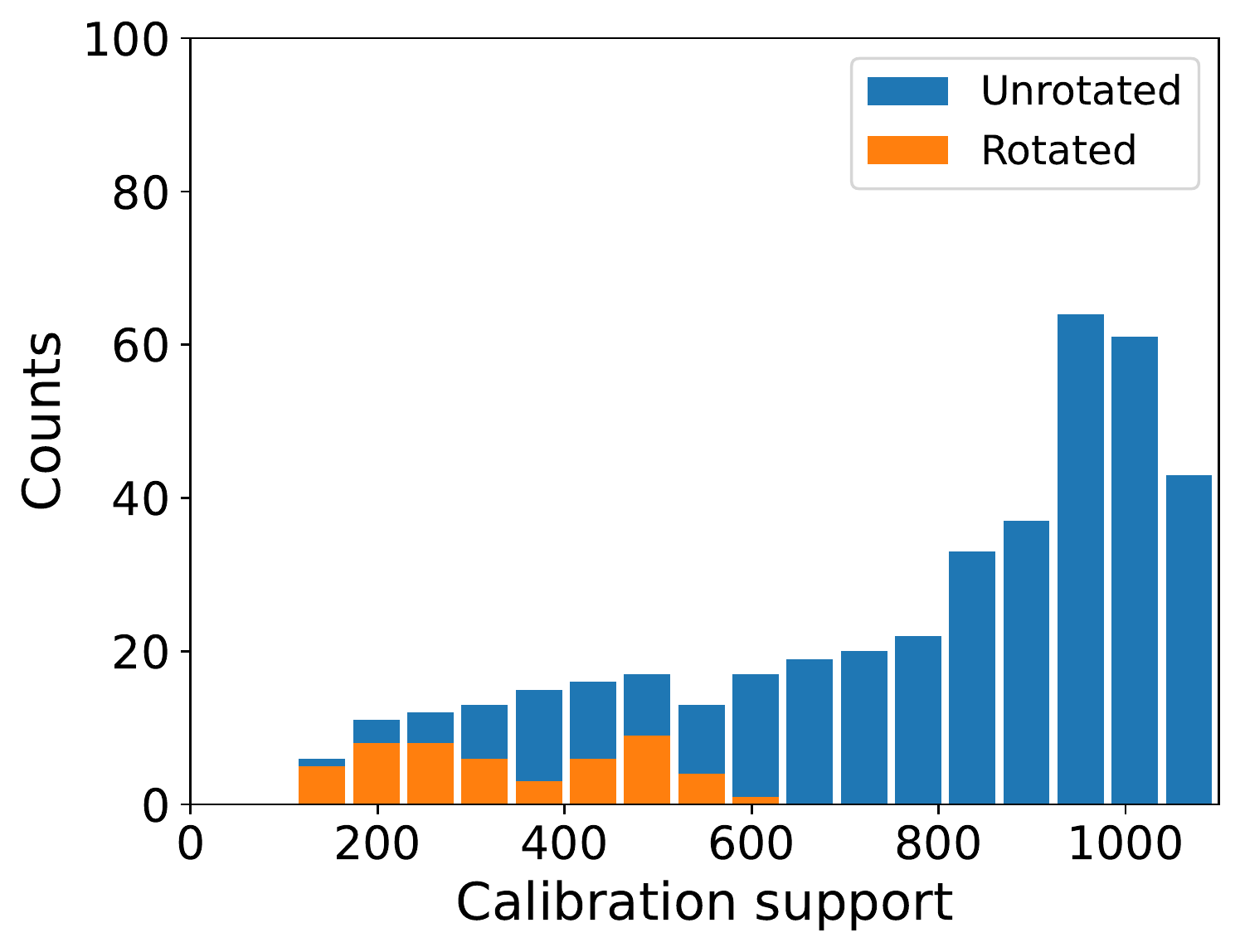}}
  \subfloat[50\% rotated]
           {\includegraphics[width=1.6in]{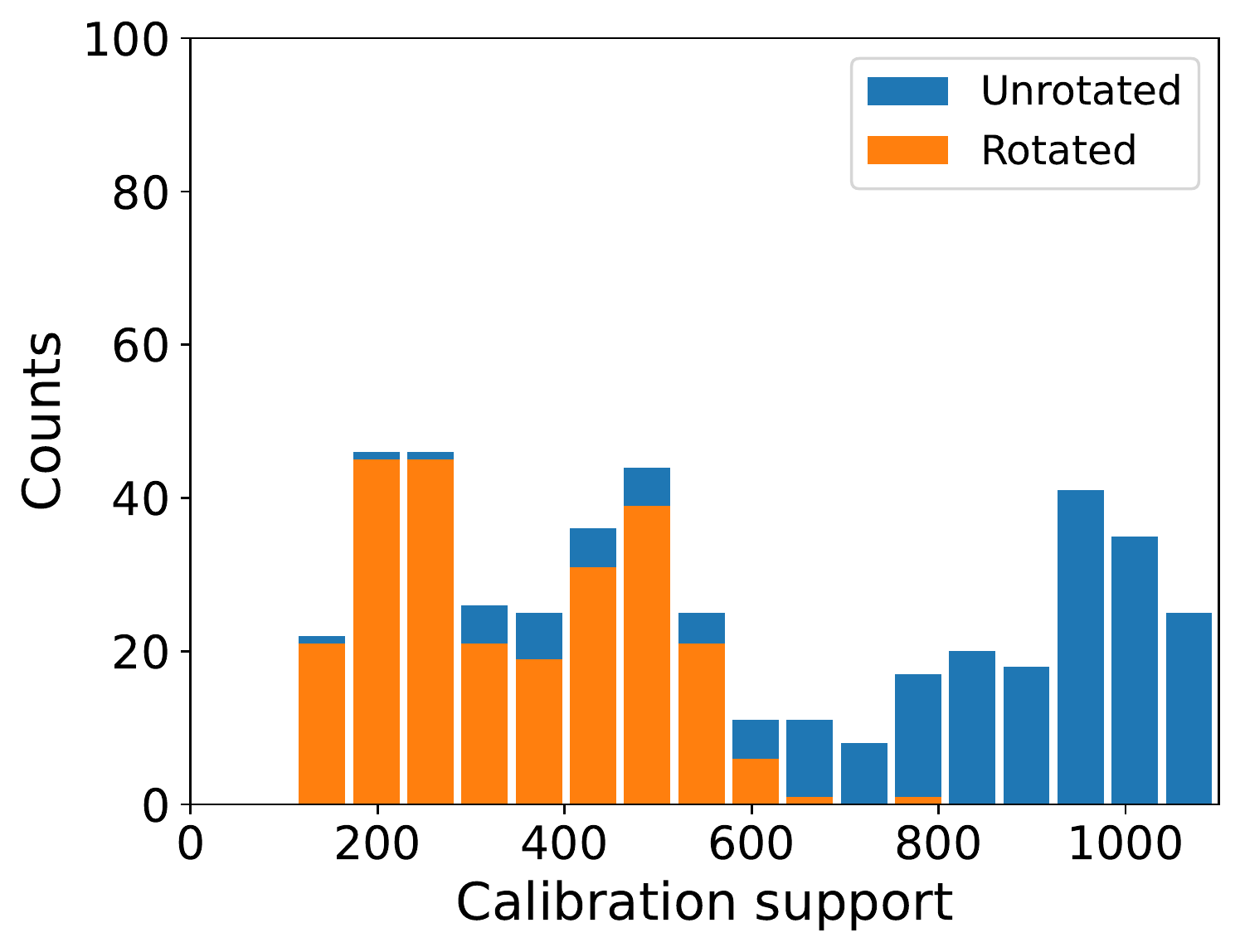}}
  \subfloat[100\% rotated]
           {\includegraphics[width=1.6in]{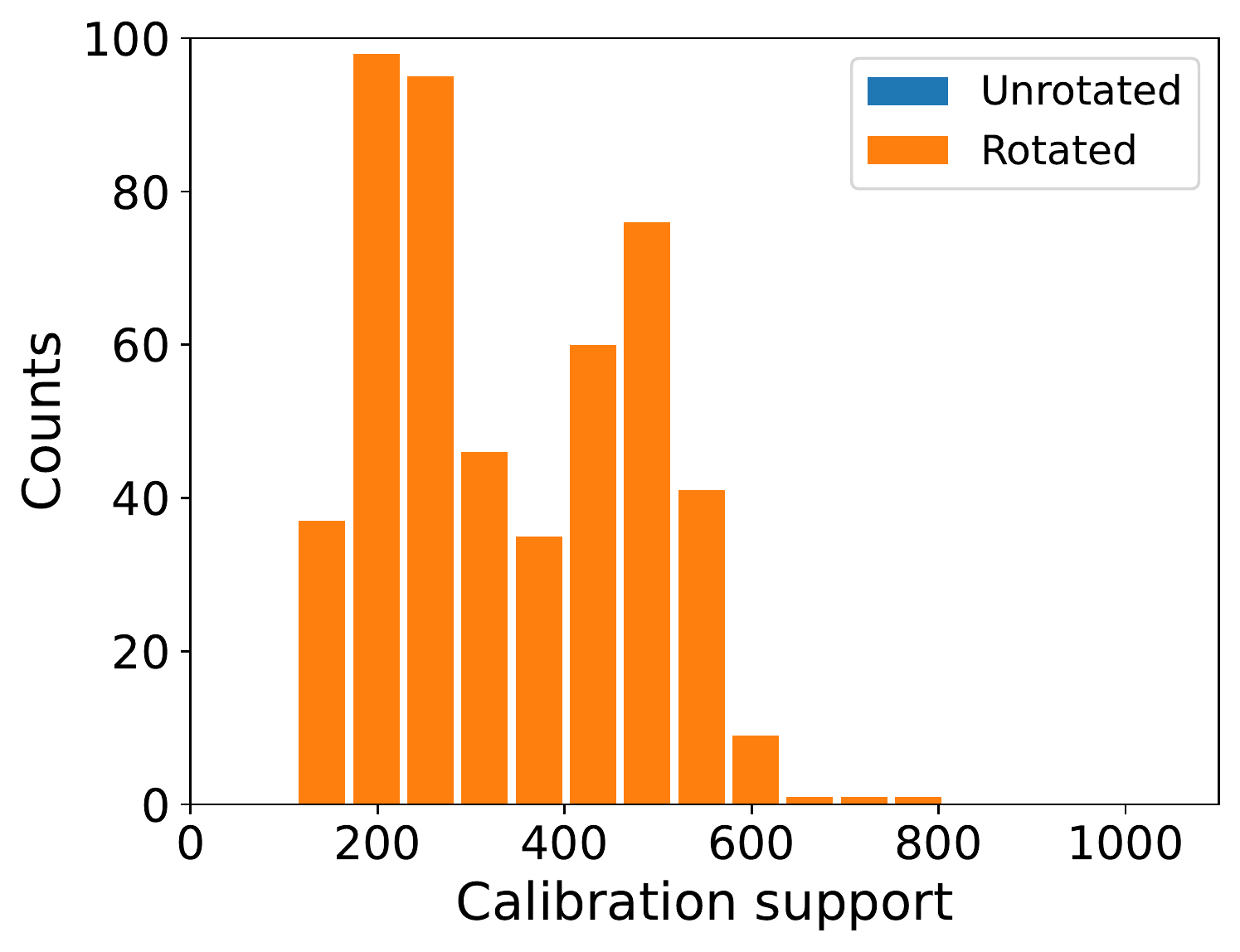}}
  \caption{Distribution of calibration support values for 500 test items
    (``mnist-1v7'') classified by a linear SVM with progressively more
    of the test set items rotated.  Bar plots are stacked.}
  \label{fig:mnist-rotate}
\end{figure}

Likewise, low calibration support can provide a warning when domain shift is
present between the calibration and test sets (or any new prediction).
We simulated domain shift (covariate shift) by rotating a 
subset of the test items 90 degrees counter-clockwise.
Figure~\ref{fig:mnist-rotate}(a) shows the distribution of
calibration support values without any rotation; the peak value is
around 900, and 
items with low support are rare.  With 10\% of the test set
rotated (Figure~\ref{fig:mnist-rotate}(b)), the overall histogram
changes a little, and the rotated items (orange bars) tend to have low
calibration support, signaling a need for inspection.
With 50\% (partial domain shift) rotated
(Figure~\ref{fig:mnist-rotate}(c)), distinct populations for the
rotated and unrotated items are clear, and with 100\% rotated
(complete domain shift), the whole histogram shifts to lower values
(peak around 300).

This result suggests that in a deployment setting, it is useful to
monitor the calibration support values that are reported by SWC.
Knowledge about typical support values (or better, their
distribution) could enable early detection of domain shift when new
items originate from a changed distribution.  That signal indicates
that the calibration set, and likely the trained model as well,
require revision.
Platt, temperature scaling, histogram binning, and other methods
provide a fixed mapping $\Phi(\hat{p})$ without regard to the item
being calibrated; there is no signal to indicate whether $\Phi$ is
still relevant.  SWC provides an intrinsic measure of
calibration set relevance through the calibration support values obtained
by each new item.


\section{Conclusions, Limitations, and Future work}
\label{sec:conc}

In this work, we explored the benefits of local, individual
calibration for each test item, in contrast to widely used global
classifier calibration methods.  We identified hidden heterogeneity
(HH) as
a strong indicator of the need for local calibration, when there are
subpopulations within a data set that have the same uncalibrated
predicted probability $\hat{p}$ yet require different corrections to
achieve a well-calibrated probability $\hat{q}$.  We provided a method
for calculating HH before calibration to inform selection of the
calibration method.  Experiments with
tabular data sets and diverse machine learning classifiers
indicate that local calibration improves Brier score in proportion to
the average hidden heterogeneity (HH) value in the data set.  We highlight
this finding as an important step towards not only correcting
miscalibration but also explaining and understanding it.
When HH is very low (as we found with several deep neural networks),
or little calibration data is available,
global methods such as temperature scaling are 
sufficient, but otherwise, local calibration is preferred.
On the other hand, because local calibration has far more
degrees of freedom than parametric, global methods,
it tends to require more calibration data.  If 
calibration data is scarce, global methods may be preferred.

We proposed a similarity-based approach to local calibration (SWC)
that weights evidence from the calibration set according to its
similarity to the test item in an ``augmented'' feature space that includes both the input features and the predicted class probabilities.  This concept
goes beyond prior work such as Similarity-Binning Averaging (SBA-10), which
calibrates (without weighting) using the 10 nearest neighbors based on
Euclidean distance in the augmented feature space~\citep{bella:sba09}.  In
most cases, we found that SWC out-performs SBA-10.  Additional benefits
can be obtained by incorporating HH directly into the SWC algorithm
(SWC-HH).  A final and unique benefit of similarity-based
calibration is that the explicit measurement of calibration support
can serve to warn when a given test item lacks good representation in
the calibration set.  This can also be an indicator when distribution
shift or domain shift is present.

\paragraph{Limitations: Runtime.}
The computational cost of local calibration methods tends to be higher
than that of global methods, since each item is independently modeled
rather than constructing a single model to apply to all items.
However, this also means that calibration can be conducted lazily, as
needed, given a similarity measure.
%
The computation of hidden heterogeneity requires
(1) the identification of an item's nearest neighbors in the probability
simplex $\Delta_{K-1}$, which can be costly with a large number of
classes, and (2) training a specialized classifier to estimate the
potential Brier score improvement (Equation~\ref{eqn:hh}).

\paragraph{Limitations: Preservation of accuracy.}
SWC and SWC-HH are not rank-preserving calibration methods.  This
means that in 
addition to modifying the calibration properties of the predictions,
they can also change the predicted class and therefore the accuracy of
predictions.  Improvements in accuracy are reflected in improved Brier
scores. 
%
Temperature scaling, in contrast, does preserve rank and
accuracy because, without a bias term, it cannot move items to the
other side of the decision threshold.  Some researchers favor
rank-preserving methods~\citep{zhang:calib20,patel:mi-calib21}, since
they seek to improve calibration without sacrificing accuracy.
However, this constraint also prevents them from {\em increasing} accuracy,
which is an outcome available to Platt scaling (via its bias term),
histogram binning, SWC, 
etc.  On the whole, we agree with~\cite{bella:sba13} that there is no
need to preserve item rankings given the opportunity to improve both
accuracy and calibration.  However, we acknowledge that in some
applications, there could 
be a need to choose a rank-preserving calibration method to ensure
accuracy is unchanged (up or down) for user
acceptance~\citep{srivastava:btc20}. 

\paragraph{Future work.}
There are several important directions for future work.
It is possible that within the same data set, some items are
best calibrated with global methods while others (where HH is present)
benefit from local calibration.  A hybrid method that selectively
applies global/local calibration, or some combination of the two, for
each test item could potentially out-perform either one.
For a given problem, alternative choices for data representation
  and similarity measures could yield additional improvements.
In addition, SWC is well designed to naturally
accommodate domain shift, if the calibration data set is drawn from
the shifted distribution.  Sampling bias in the training set, whether
intentional or not, induces a particular kind of domain shift that is
especially important to address to meet fairness goals when
predictions are made in a deployment setting.
%


\subsubsection*{Author Contributions}
KW: Problem refinement, algorithm development, implementation, experimentation, data analysis, and writing. \\
TD: Initial problem formulation, critical feedback, assisting with writing.

\subsubsection*{Acknowledgments}
This material is based upon work supported by the Defense Advanced
Research Projects Agency (DARPA) under Contract No.~HR001119C0112. Any
opinions, findings and conclusions or recommendations expressed in
this material are those of the authors and do not necessarily reflect
the views of the DARPA. The authors thank the reviewers and the editor for valuable discussions that significantly improved the paper.

\bibliography{calib}
\bibliographystyle{tmlr}

\newpage
\appendix
\section{Appendix}
\label{app:res}

\begin{figure}
  \centering
  \subfloat[mnist-1v7 Brier score]{\includegraphics[width=2.89in]
    {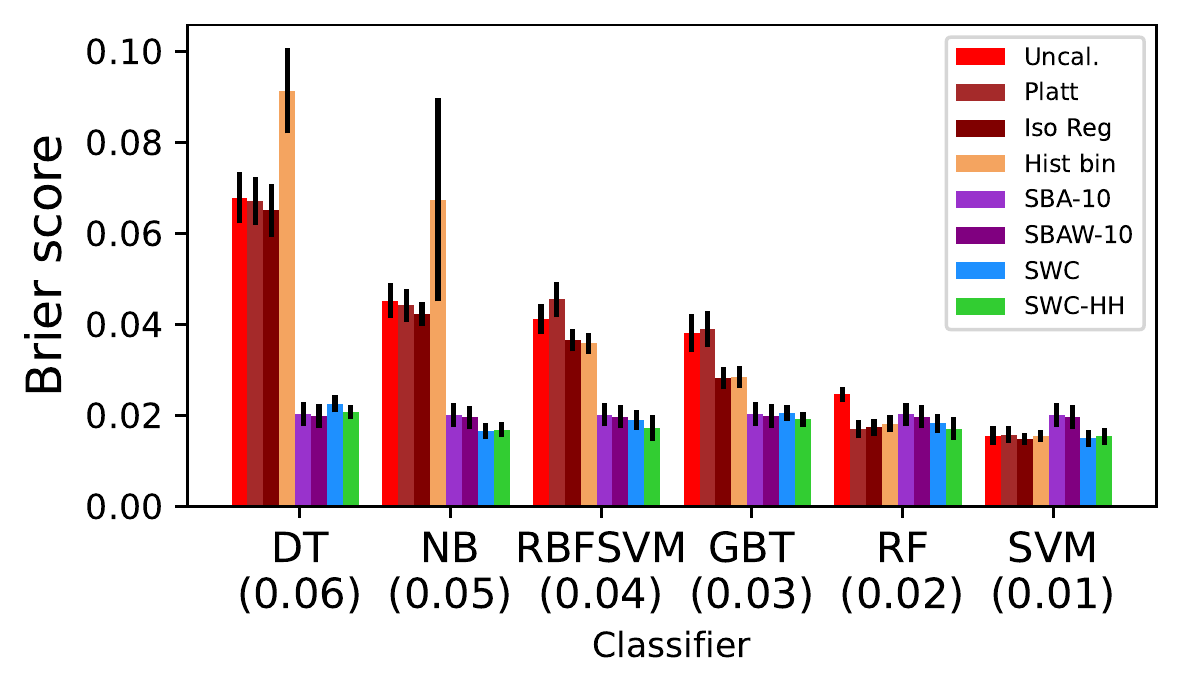}}
  \subfloat[mnist-1v7 accuracy]{\includegraphics[width=2.89in]
    {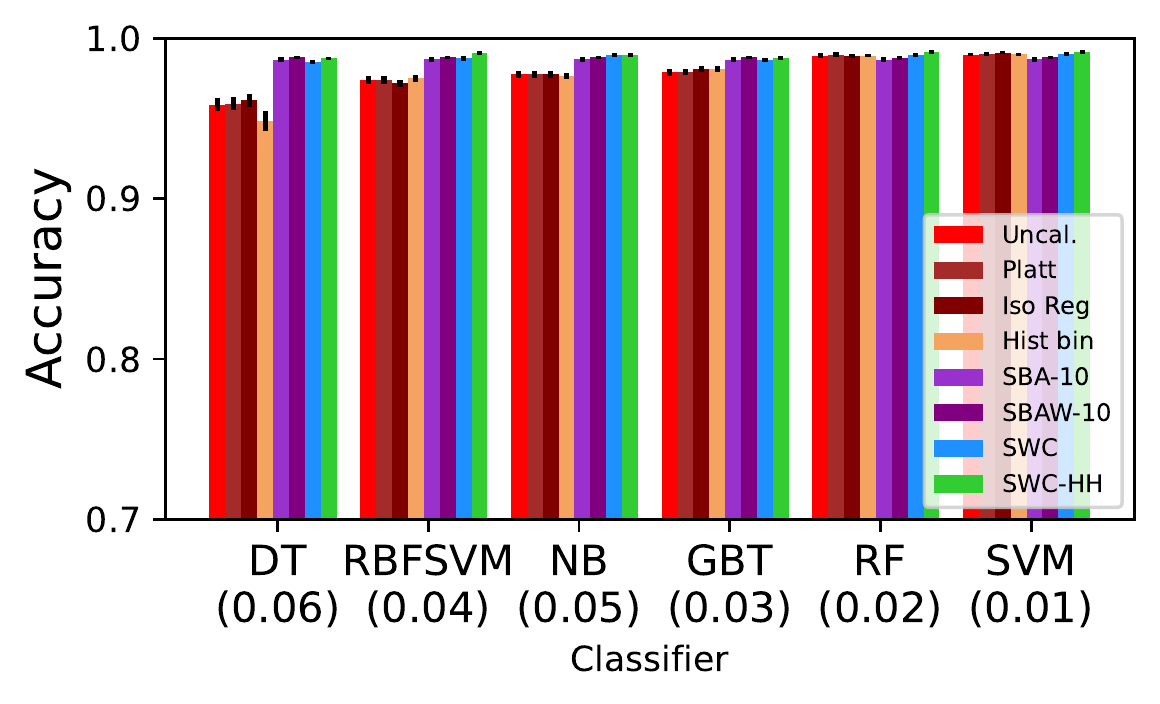}}
  \\
  \subfloat[mnist-4v9 Brier score]{\includegraphics[width=2.89in]
    {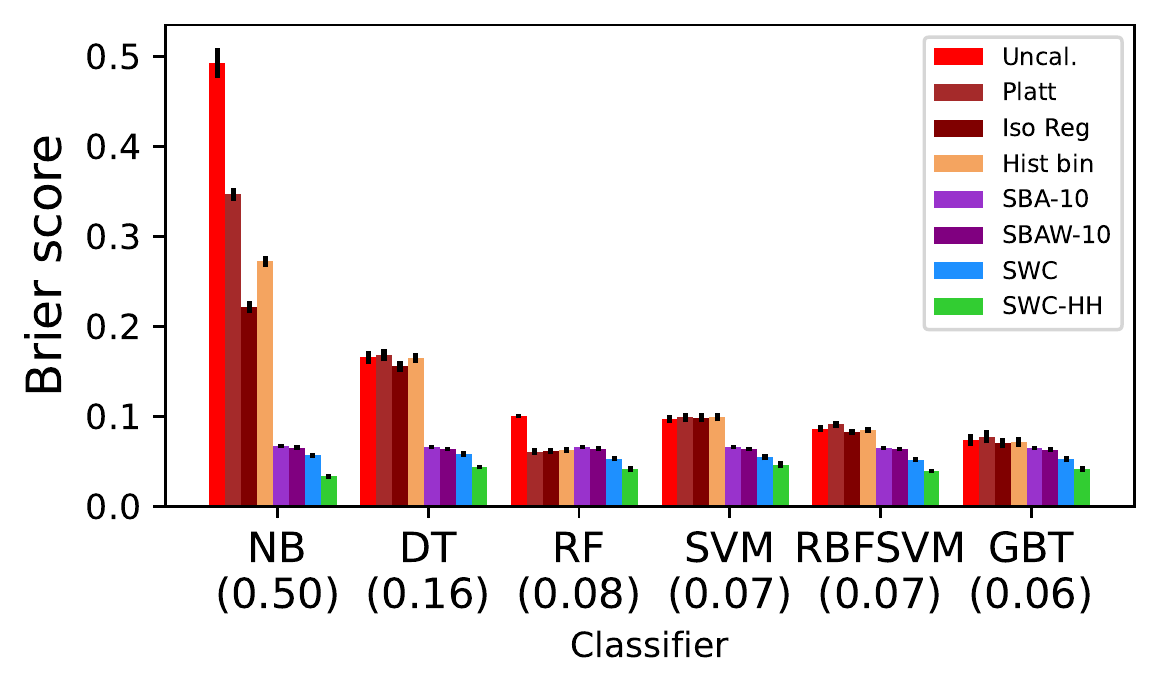}}
  \subfloat[mnist-4v9 accuracy]{\includegraphics[width=2.89in]
    {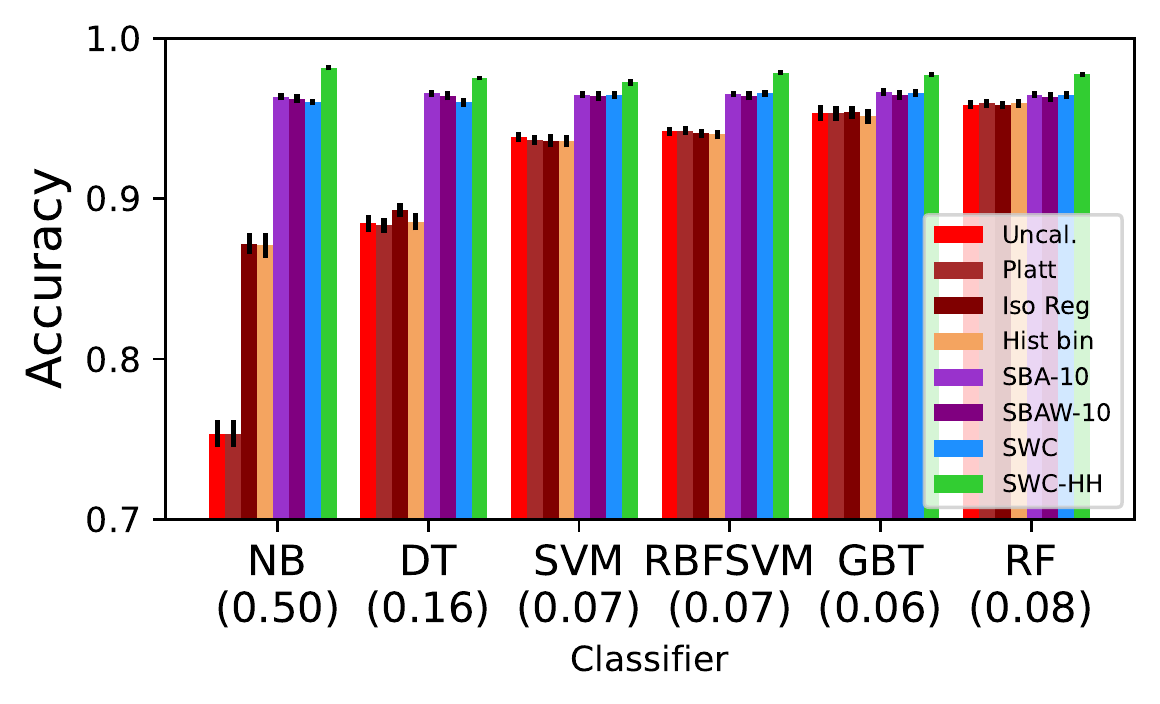}}
  \\
  \subfloat[mnist-3v8 Brier score]{\includegraphics[width=2.89in]
    {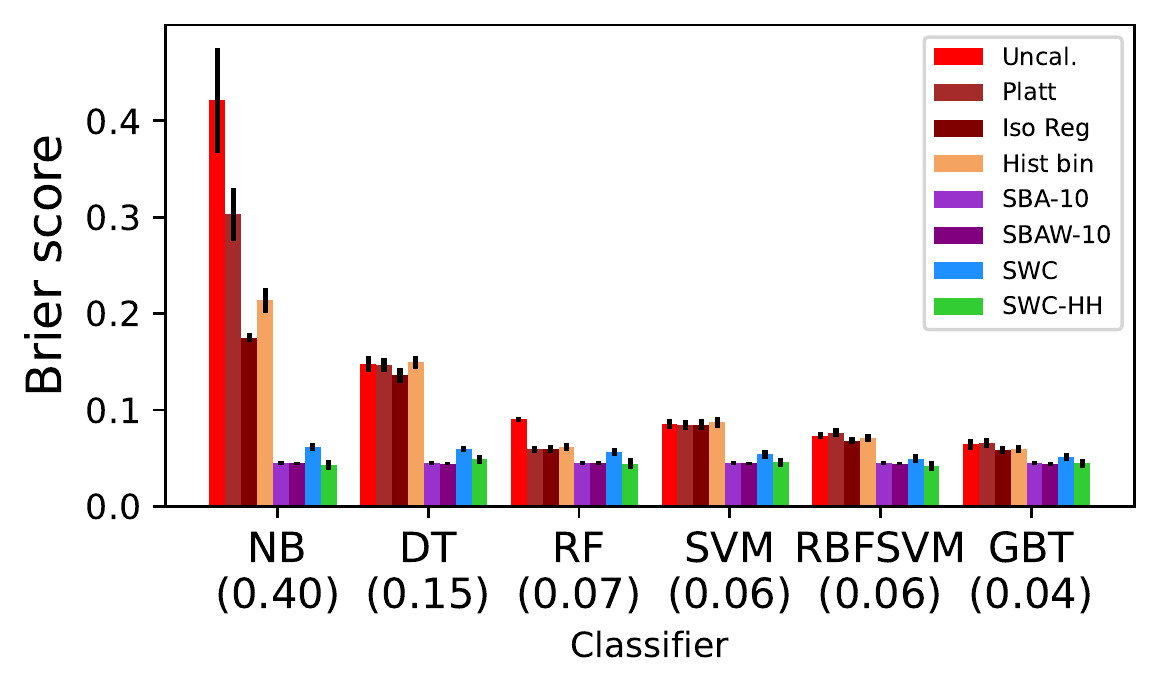}}
  \subfloat[mnist-3v8 accuracy]{\includegraphics[width=2.89in]
    {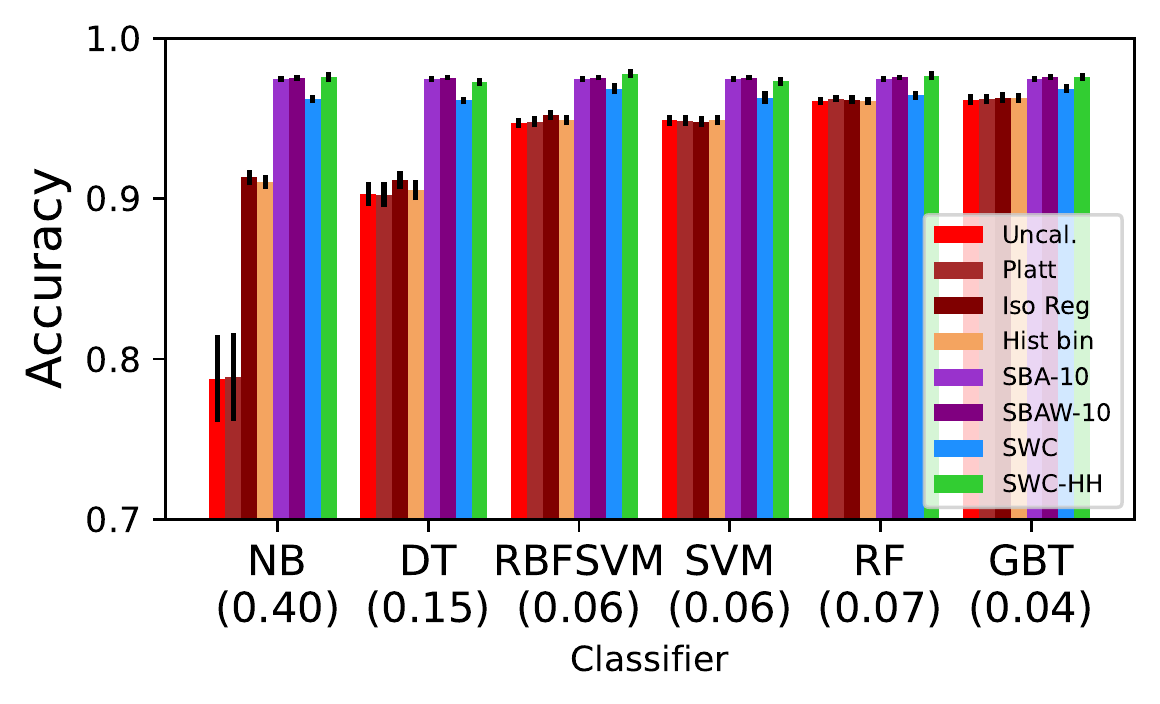}}
  \\
  \subfloat[mnist-3v5 Brier score]{\includegraphics[width=2.89in]
    {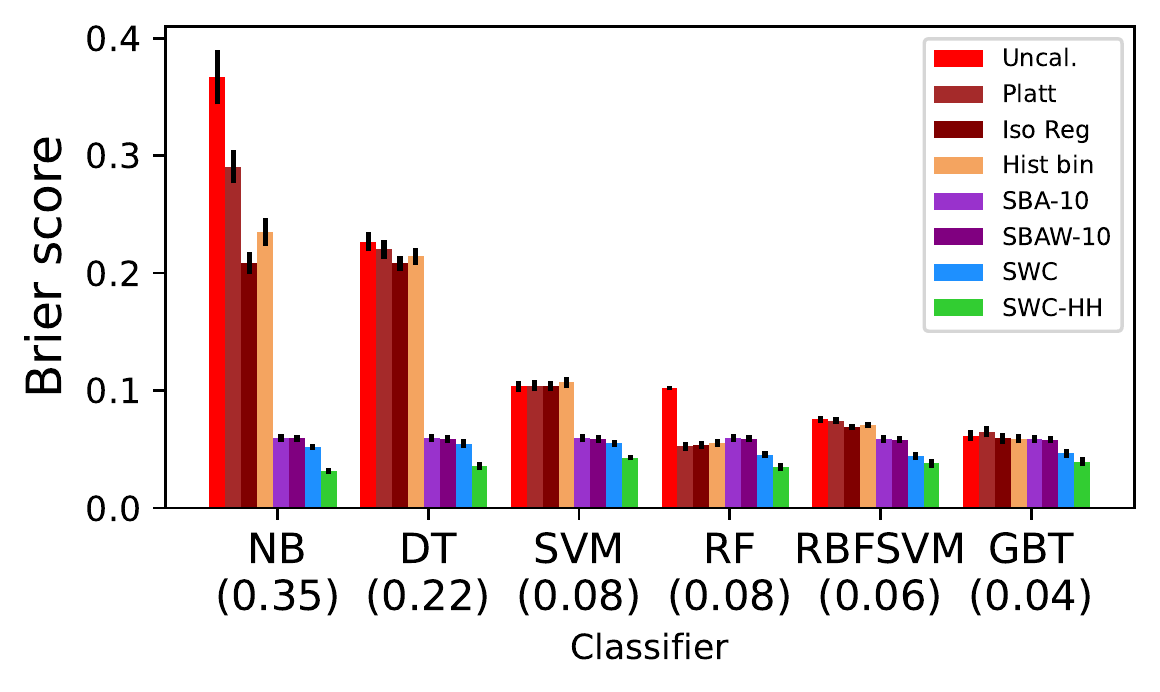}}
  \subfloat[mnist-3v5 accuracy]{\includegraphics[width=2.89in]
    {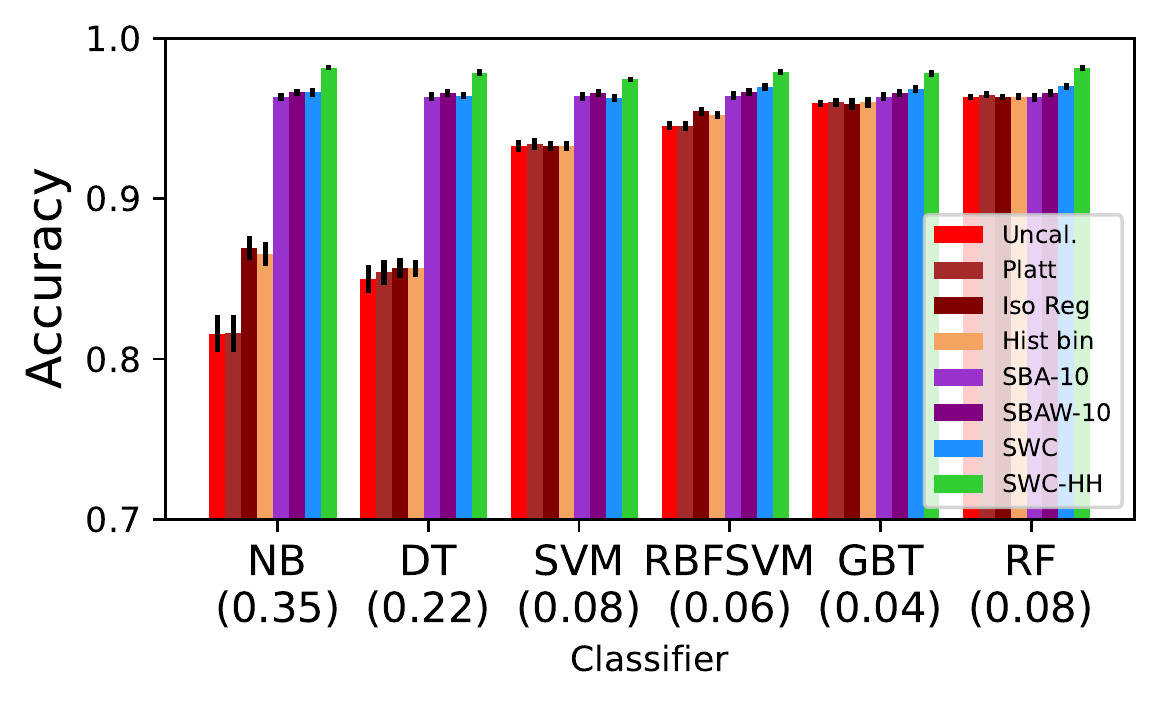}}
  \caption{Calibration performance (left) and accuracy (right)
  for the binary MNIST data sets
  (10 trials; error bars indicate one standard error). 
  Classifiers are sorted in order of improvement based on the
  uncalibrated classifier's score, and average HH values are below
  each classifier.}
    \label{fig:res-tabular-binary}
\end{figure}

This appendix provides experimental results for seven data sets and
six classifiers, comparing similarity-based calibration to other
methods.  In these experiments, we randomly sampled \num{10000} items
from each data set and randomly split them into \num{500} train (for
binary problems) or \num{1000} train (for multi-class problems),
\num{500} test, and \num{9000} for a calibration pool.

\subsection{Binary tabular data}
Figure~\ref{fig:res-tabular-binary}
presents results across all classifiers and calibration methods
for the four binary MNIST data sets after \num{3000} calibration items were
employed.  Complete numeric results are shown in
  Tables~\ref{tab:mnist-1v7} to~\ref{tab:mnist-3v5}. Classifiers appear in
order of improving (decreasing) Brier score for the uncalibrated predictions.

Platt scaling improved performance for the Naive Bayes and random
forest classifiers, but it yielded little benefit for the others.
Isotonic regression sometimes provided additional improvements,
  primarily for Naive Bayes, and usually out-performed histogram binning.
In contrast, similarity-based methods were
highly effective for all classifiers in reducing Brier score.

SWC-HH consistently provided the best results.  
The SBA-10 approach described by~\cite{bella:sba09} weights all ten
neighbors equally.  We learned (Ferri, personal communication,
Oct.~29, 2022) that the SBA authors have subsequently employed
weighting in the averaging process so that each neighbor contributes
inversely to its distance from the item to be calibrated.
We included the weighted variation in our experiments, denoted as
``SBAW-10''.  Weighting provides slight advantages over SBA-10 in some
cases, but SWC/SWC-HH yielded the best results.  We can view SBAW as
an intermediate choice between SBA and SWC, as  it adopts the distance
weighting of SWC but not the other aspects (supervised distance
metric, HH filtering) that make SWC-HH the strongest method overall.
As noted earlier, and most evident in Tables~\ref{tab:mnist-1v7}
to~\ref{tab:mnist-3v5}, both SBA and SBAW generate nearly identical
results for all classifiers, because they are dominated by feature
space distance, and the classifier's initial predictions have little
influence.  In contrast, SWC and SWC-HH adapt to each classifier's
individual limitations (hidden heterogeneity).  
SWC/SWC-HH improvements correlate with the average HH
value, shown in parenthesis under the x axis labels.
In addition, similarity-based calibration also increased test
accuracy (see right column of Figure~\ref{fig:res-tabular-binary}
and subtables in Tables~\ref{tab:mnist-1v7} to~\ref{tab:mnist-3v5}).
SWC-HH consistently achieved the highest accuracy.

\setlength{\tabcolsep}{5pt}
\begin{table}[ht]
  \caption{Results for mnist-1v7
    ($n_{cal}=3000$, 10 trials).
    The best result(s) for each model (within 1 standard error, shown as a subscript) are in bold.}
  \label{tab:mnist-1v7}
  \small
  \begin{tabular}{l|c|ccc|cccc} 
  \hline
  \multicolumn{9}{c}{Brier score}\\ \hline
    Model & Uncal. & Platt & Iso Reg & Hist bin & SBA-10 & SBAW-10 & SWC & SWC-HH \\ \hline
    \footnotesize{DT} & \footnotesize{0.0678}$_{0.006}$ & \footnotesize{0.0671}$_{0.005}$ & \footnotesize{0.0650}$_{0.006}$ & \footnotesize{0.0913}$_{0.009}$ & \footnotesize{\bf 0.0203}$_{0.003}$ & \footnotesize{\bf 0.0198}$_{0.003}$ & \footnotesize{0.0226}$_{0.002}$ & \footnotesize{\bf 0.0207}$_{0.002}$ \\
    \footnotesize{NB} & \footnotesize{0.0452}$_{0.004}$ & \footnotesize{0.0442}$_{0.004}$ & \footnotesize{0.0423}$_{0.003}$ & \footnotesize{0.0674}$_{0.022}$ & \footnotesize{0.0200}$_{0.003}$ & \footnotesize{0.0195}$_{0.003}$ & \footnotesize{\bf 0.0165}$_{0.002}$ & \footnotesize{\bf 0.0168}$_{0.002}$ \\
    \footnotesize{RBFSVM} & \footnotesize{0.0411}$_{0.003}$ & \footnotesize{0.0455}$_{0.004}$ & \footnotesize{0.0365}$_{0.002}$ & \footnotesize{0.0358}$_{0.002}$ & \footnotesize{0.0201}$_{0.003}$ & \footnotesize{\bf 0.0197}$_{0.003}$ & \footnotesize{\bf 0.0189}$_{0.002}$ & \footnotesize{\bf 0.0172}$_{0.003}$ \\
    \footnotesize{GBT} & \footnotesize{0.0380}$_{0.004}$ & \footnotesize{0.0390}$_{0.004}$ & \footnotesize{0.0282}$_{0.003}$ & \footnotesize{0.0284}$_{0.002}$ & \footnotesize{\bf 0.0203}$_{0.003}$ & \footnotesize{\bf 0.0198}$_{0.003}$ & \footnotesize{\bf 0.0205}$_{0.002}$ & \footnotesize{\bf 0.0191}$_{0.002}$ \\
    \footnotesize{RF} & \footnotesize{0.0246}$_{0.002}$ & \footnotesize{\bf 0.0170}$_{0.002}$ & \footnotesize{\bf 0.0173}$_{0.002}$ & \footnotesize{\bf 0.0182}$_{0.002}$ & \footnotesize{0.0202}$_{0.003}$ & \footnotesize{0.0197}$_{0.003}$ & \footnotesize{\bf 0.0182}$_{0.002}$ & \footnotesize{\bf 0.0171}$_{0.002}$ \\
    \footnotesize{SVM} & \footnotesize{\bf 0.0155}$_{0.002}$ & \footnotesize{\bf 0.0157}$_{0.002}$ & \footnotesize{\bf 0.0148}$_{0.001}$ & \footnotesize{\bf 0.0155}$_{0.001}$ & \footnotesize{0.0200}$_{0.003}$ & \footnotesize{0.0196}$_{0.003}$ & \footnotesize{\bf 0.0149}$_{0.002}$ & \footnotesize{\bf 0.0154}$_{0.002}$ \\
  \hline
  \multicolumn{9}{c}{Accuracy}\\ \hline
    Model & Uncal. & Platt & Iso Reg & Hist bin & SBA-10 & SBAW-10 & SWC & SWC-HH \\ \hline
    \footnotesize{DT} & \footnotesize{0.9586}$_{0.004}$ & \footnotesize{0.9592}$_{0.004}$ & \footnotesize{0.9614}$_{0.004}$ & \footnotesize{0.9484}$_{0.006}$ & \footnotesize{\bf 0.9866}$_{0.002}$ & \footnotesize{\bf 0.9880}$_{0.001}$ & \footnotesize{0.9852}$_{0.001}$ & \footnotesize{\bf 0.9874}$_{0.001}$ \\
    \footnotesize{RBFSVM} & \footnotesize{0.9740}$_{0.003}$ & \footnotesize{0.9738}$_{0.002}$ & \footnotesize{0.9718}$_{0.002}$ & \footnotesize{0.9750}$_{0.002}$ & \footnotesize{0.9868}$_{0.001}$ & \footnotesize{0.9880}$_{0.001}$ & \footnotesize{0.9876}$_{0.002}$ & \footnotesize{\bf 0.9906}$_{0.001}$ \\
    \footnotesize{NB} & \footnotesize{0.9774}$_{0.002}$ & \footnotesize{0.9774}$_{0.002}$ & \footnotesize{0.9774}$_{0.002}$ & \footnotesize{0.9766}$_{0.002}$ & \footnotesize{0.9868}$_{0.001}$ & \footnotesize{0.9880}$_{0.001}$ & \footnotesize{\bf 0.9894}$_{0.001}$ & \footnotesize{\bf 0.9894}$_{0.001}$ \\
    \footnotesize{GBT} & \footnotesize{0.9788}$_{0.002}$ & \footnotesize{0.9790}$_{0.002}$ & \footnotesize{0.9808}$_{0.002}$ & \footnotesize{0.9808}$_{0.002}$ & \footnotesize{\bf 0.9866}$_{0.002}$ & \footnotesize{\bf 0.9880}$_{0.001}$ & \footnotesize{0.9864}$_{0.001}$ & \footnotesize{\bf 0.9878}$_{0.001}$ \\
    \footnotesize{RF} & \footnotesize{0.9890}$_{0.001}$ & \footnotesize{\bf 0.9898}$_{0.001}$ & \footnotesize{0.9888}$_{0.001}$ & \footnotesize{0.9892}$_{0.001}$ & \footnotesize{0.9866}$_{0.001}$ & \footnotesize{0.9878}$_{0.001}$ & \footnotesize{0.9894}$_{0.001}$ & \footnotesize{\bf 0.9912}$_{0.001}$ \\
    \footnotesize{SVM} & \footnotesize{0.9898}$_{0.001}$ & \footnotesize{\bf 0.9904}$_{0.001}$ & \footnotesize{\bf 0.9910}$_{0.001}$ & \footnotesize{0.9900}$_{0.001}$ & \footnotesize{0.9868}$_{0.001}$ & \footnotesize{0.9880}$_{0.001}$ & \footnotesize{\bf 0.9902}$_{0.001}$ & \footnotesize{\bf 0.9914}$_{0.001}$ \\
    \hline
  \end{tabular}
\end{table}

\begin{table}[ht]
  \caption{Results for mnist-4v9
    ($n_{cal}=3000$, 10 trials).
    The best result(s) for each model (within 1 standard error, shown as a subscript) are in bold.}
  \label{tab:mnist-4v9}
  \small
  \begin{tabular}{l|c|ccc|cccc} 
  \hline
  \multicolumn{9}{c}{Brier score}\\ \hline
    Model & Uncal. & Platt & Iso Reg & Hist bin & SBA-10 & SBAW-10 & SWC & SWC-HH \\ \hline
    \footnotesize{NB} & \footnotesize{0.4932}$_{0.017}$ & \footnotesize{0.3472}$_{0.007}$ & \footnotesize{0.2216}$_{0.007}$ & \footnotesize{0.2724}$_{0.006}$ & \footnotesize{0.0671}$_{0.002}$ & \footnotesize{0.0653}$_{0.002}$ & \footnotesize{0.0565}$_{0.003}$ & \footnotesize{\bf 0.0335}$_{0.003}$ \\
    \footnotesize{DT} & \footnotesize{0.1657}$_{0.007}$ & \footnotesize{0.1679}$_{0.007}$ & \footnotesize{0.1558}$_{0.006}$ & \footnotesize{0.1651}$_{0.006}$ & \footnotesize{0.0655}$_{0.002}$ & \footnotesize{0.0638}$_{0.002}$ & \footnotesize{0.0585}$_{0.003}$ & \footnotesize{\bf 0.0436}$_{0.003}$ \\
    \footnotesize{RF} & \footnotesize{0.1005}$_{0.002}$ & \footnotesize{0.0609}$_{0.003}$ & \footnotesize{0.0619}$_{0.003}$ & \footnotesize{0.0627}$_{0.004}$ & \footnotesize{0.0659}$_{0.002}$ & \footnotesize{0.0642}$_{0.002}$ & \footnotesize{0.0528}$_{0.003}$ & \footnotesize{\bf 0.0410}$_{0.003}$ \\
    \footnotesize{SVM} & \footnotesize{0.0975}$_{0.004}$ & \footnotesize{0.0988}$_{0.005}$ & \footnotesize{0.0984}$_{0.005}$ & \footnotesize{0.0993}$_{0.005}$ & \footnotesize{0.0655}$_{0.002}$ & \footnotesize{0.0638}$_{0.002}$ & \footnotesize{0.0547}$_{0.004}$ & \footnotesize{\bf 0.0462}$_{0.004}$ \\
    \footnotesize{RBFSVM} & \footnotesize{0.0863}$_{0.004}$ & \footnotesize{0.0912}$_{0.004}$ & \footnotesize{0.0827}$_{0.004}$ & \footnotesize{0.0851}$_{0.004}$ & \footnotesize{0.0652}$_{0.002}$ & \footnotesize{0.0636}$_{0.002}$ & \footnotesize{0.0518}$_{0.003}$ & \footnotesize{\bf 0.0393}$_{0.002}$ \\
    \footnotesize{GBT} & \footnotesize{0.0741}$_{0.007}$ & \footnotesize{0.0773}$_{0.007}$ & \footnotesize{0.0702}$_{0.006}$ & \footnotesize{0.0717}$_{0.006}$ & \footnotesize{0.0647}$_{0.002}$ & \footnotesize{0.0630}$_{0.002}$ & \footnotesize{0.0522}$_{0.003}$ & \footnotesize{\bf 0.0413}$_{0.003}$ \\
  \hline
  \multicolumn{9}{c}{Accuracy}\\ \hline
    Model & Uncal. & Platt & Iso Reg & Hist bin & SBA-10 & SBAW-10 & SWC & SWC-HH \\ \hline
    \footnotesize{NB} & \footnotesize{0.7534}$_{0.008}$ & \footnotesize{0.7534}$_{0.008}$ & \footnotesize{0.8720}$_{0.007}$ & \footnotesize{0.8710}$_{0.008}$ & \footnotesize{0.9634}$_{0.002}$ & \footnotesize{0.9624}$_{0.003}$ & \footnotesize{0.9602}$_{0.002}$ & \footnotesize{\bf 0.9816}$_{0.002}$ \\
    \footnotesize{DT} & \footnotesize{0.8846}$_{0.005}$ & \footnotesize{0.8834}$_{0.005}$ & \footnotesize{0.8928}$_{0.005}$ & \footnotesize{0.8856}$_{0.005}$ & \footnotesize{0.9656}$_{0.002}$ & \footnotesize{0.9642}$_{0.003}$ & \footnotesize{0.9600}$_{0.003}$ & \footnotesize{\bf 0.9752}$_{0.001}$ \\
    \footnotesize{SVM} & \footnotesize{0.9382}$_{0.003}$ & \footnotesize{0.9368}$_{0.003}$ & \footnotesize{0.9360}$_{0.004}$ & \footnotesize{0.9358}$_{0.004}$ & \footnotesize{0.9648}$_{0.002}$ & \footnotesize{0.9638}$_{0.003}$ & \footnotesize{0.9646}$_{0.002}$ & \footnotesize{\bf 0.9726}$_{0.002}$ \\
    \footnotesize{RBFSVM} & \footnotesize{0.9420}$_{0.003}$ & \footnotesize{0.9424}$_{0.003}$ & \footnotesize{0.9406}$_{0.003}$ & \footnotesize{0.9400}$_{0.003}$ & \footnotesize{0.9652}$_{0.002}$ & \footnotesize{0.9642}$_{0.003}$ & \footnotesize{0.9656}$_{0.002}$ & \footnotesize{\bf 0.9786}$_{0.001}$ \\
    \footnotesize{GBT} & \footnotesize{0.9534}$_{0.005}$ & \footnotesize{0.9532}$_{0.005}$ & \footnotesize{0.9538}$_{0.004}$ & \footnotesize{0.9514}$_{0.005}$ & \footnotesize{0.9664}$_{0.002}$ & \footnotesize{0.9646}$_{0.003}$ & \footnotesize{0.9656}$_{0.002}$ & \footnotesize{\bf 0.9772}$_{0.002}$ \\
    \footnotesize{RF} & \footnotesize{0.9586}$_{0.003}$ & \footnotesize{0.9594}$_{0.003}$ & \footnotesize{0.9584}$_{0.003}$ & \footnotesize{0.9594}$_{0.003}$ & \footnotesize{0.9648}$_{0.002}$ & \footnotesize{0.9632}$_{0.003}$ & \footnotesize{0.9644}$_{0.003}$ & \footnotesize{\bf 0.9774}$_{0.002}$ \\
    \hline
  \end{tabular}
\end{table}

\begin{table}[ht]
  \caption{Results for mnist-3v8
    ($n_{cal}=3000$, 10 trials).
    The best result(s) for each model (within 1 standard error, shown as a subscript) are in bold.}
  \label{tab:mnist-3v8}
  \small
  \begin{tabular}{l|c|ccc|cccc} 
  \hline
  \multicolumn{9}{c}{Brier score}\\ \hline
    Model & Uncal. & Platt & Iso Reg & Hist bin & SBA-10 & SBAW-10 & SWC & SWC-HH \\ \hline
    \footnotesize{NB} & \footnotesize{0.4212}$_{0.054}$ & \footnotesize{0.3029}$_{0.028}$ & \footnotesize{0.1749}$_{0.004}$ & \footnotesize{0.2136}$_{0.013}$ & \footnotesize{0.0451}$_{0.002}$ & \footnotesize{\bf 0.0446}$_{0.002}$ & \footnotesize{0.0613}$_{0.004}$ & \footnotesize{\bf 0.0429}$_{0.005}$ \\
    \footnotesize{DT} & \footnotesize{0.1477}$_{0.008}$ & \footnotesize{0.1465}$_{0.008}$ & \footnotesize{0.1359}$_{0.008}$ & \footnotesize{0.1494}$_{0.007}$ & \footnotesize{\bf 0.0448}$_{0.002}$ & \footnotesize{\bf 0.0443}$_{0.002}$ & \footnotesize{0.0594}$_{0.004}$ & \footnotesize{\bf 0.0486}$_{0.004}$ \\
    \footnotesize{RF} & \footnotesize{0.0902}$_{0.003}$ & \footnotesize{0.0590}$_{0.004}$ & \footnotesize{0.0595}$_{0.004}$ & \footnotesize{0.0614}$_{0.004}$ & \footnotesize{\bf 0.0452}$_{0.002}$ & \footnotesize{\bf 0.0446}$_{0.002}$ & \footnotesize{0.0559}$_{0.004}$ & \footnotesize{\bf 0.0441}$_{0.006}$ \\
    \footnotesize{SVM} & \footnotesize{0.0853}$_{0.005}$ & \footnotesize{0.0848}$_{0.005}$ & \footnotesize{0.0847}$_{0.005}$ & \footnotesize{0.0871}$_{0.005}$ & \footnotesize{\bf 0.0450}$_{0.002}$ & \footnotesize{\bf 0.0444}$_{0.002}$ & \footnotesize{0.0539}$_{0.005}$ & \footnotesize{\bf 0.0454}$_{0.004}$ \\
    \footnotesize{RBFSVM} & \footnotesize{0.0733}$_{0.004}$ & \footnotesize{0.0764}$_{0.005}$ & \footnotesize{0.0682}$_{0.004}$ & \footnotesize{0.0707}$_{0.004}$ & \footnotesize{0.0448}$_{0.002}$ & \footnotesize{0.0443}$_{0.002}$ & \footnotesize{0.0494}$_{0.005}$ & \footnotesize{\bf 0.0418}$_{0.005}$ \\
    \footnotesize{GBT} & \footnotesize{0.0646}$_{0.006}$ & \footnotesize{0.0657}$_{0.005}$ & \footnotesize{0.0581}$_{0.004}$ & \footnotesize{0.0593}$_{0.004}$ & \footnotesize{\bf 0.0446}$_{0.002}$ & \footnotesize{\bf 0.0441}$_{0.002}$ & \footnotesize{0.0508}$_{0.004}$ & \footnotesize{\bf 0.0445}$_{0.005}$ \\
  \hline
  \multicolumn{9}{c}{Accuracy}\\ \hline
    Model & Uncal. & Platt & Iso Reg & Hist bin & SBA-10 & SBAW-10 & SWC & SWC-HH \\ \hline
    \footnotesize{NB} & \footnotesize{0.7878}$_{0.027}$ & \footnotesize{0.7888}$_{0.027}$ & \footnotesize{0.9132}$_{0.005}$ & \footnotesize{0.9104}$_{0.004}$ & \footnotesize{\bf 0.9744}$_{0.002}$ & \footnotesize{\bf 0.9750}$_{0.002}$ & \footnotesize{0.9620}$_{0.003}$ & \footnotesize{\bf 0.9760}$_{0.003}$ \\
    \footnotesize{DT} & \footnotesize{0.9030}$_{0.008}$ & \footnotesize{0.9024}$_{0.008}$ & \footnotesize{0.9116}$_{0.006}$ & \footnotesize{0.9054}$_{0.006}$ & \footnotesize{\bf 0.9744}$_{0.002}$ & \footnotesize{\bf 0.9754}$_{0.001}$ & \footnotesize{0.9612}$_{0.002}$ & \footnotesize{0.9726}$_{0.002}$ \\
    \footnotesize{RBFSVM} & \footnotesize{0.9472}$_{0.003}$ & \footnotesize{0.9480}$_{0.004}$ & \footnotesize{0.9522}$_{0.003}$ & \footnotesize{0.9488}$_{0.003}$ & \footnotesize{0.9744}$_{0.002}$ & \footnotesize{0.9754}$_{0.002}$ & \footnotesize{0.9684}$_{0.003}$ & \footnotesize{\bf 0.9778}$_{0.003}$ \\
    \footnotesize{SVM} & \footnotesize{0.9488}$_{0.003}$ & \footnotesize{0.9486}$_{0.003}$ & \footnotesize{0.9480}$_{0.004}$ & \footnotesize{0.9490}$_{0.003}$ & \footnotesize{\bf 0.9744}$_{0.002}$ & \footnotesize{\bf 0.9754}$_{0.002}$ & \footnotesize{0.9630}$_{0.004}$ & \footnotesize{\bf 0.9732}$_{0.003}$ \\
    \footnotesize{RF} & \footnotesize{0.9608}$_{0.003}$ & \footnotesize{0.9624}$_{0.002}$ & \footnotesize{0.9616}$_{0.003}$ & \footnotesize{0.9608}$_{0.003}$ & \footnotesize{0.9744}$_{0.002}$ & \footnotesize{\bf 0.9756}$_{0.001}$ & \footnotesize{0.9644}$_{0.003}$ & \footnotesize{\bf 0.9766}$_{0.003}$ \\
    \footnotesize{GBT} & \footnotesize{0.9616}$_{0.004}$ & \footnotesize{0.9620}$_{0.003}$ & \footnotesize{0.9630}$_{0.003}$ & \footnotesize{0.9628}$_{0.003}$ & \footnotesize{\bf 0.9744}$_{0.002}$ & \footnotesize{\bf 0.9758}$_{0.002}$ & \footnotesize{0.9686}$_{0.003}$ & \footnotesize{\bf 0.9758}$_{0.002}$ \\
    \hline
  \end{tabular}
\end{table}

\begin{table}[ht]
  \caption{Results for mnist-3v5
    ($n_{cal}=3000$, 10 trials).
    The best result(s) for each model (within 1 standard error, shown as a subscript) are in bold.}
  \label{tab:mnist-3v5}
  \small
  \begin{tabular}{l|c|ccc|cccc} 
  \hline
  \multicolumn{9}{c}{Brier score}\\ \hline
    Model & Uncal. & Platt & Iso Reg & Hist bin & SBA-10 & SBAW-10 & SWC & SWC-HH \\ \hline
    \footnotesize{NB} & \footnotesize{0.3670}$_{0.023}$ & \footnotesize{0.2905}$_{0.014}$ & \footnotesize{0.2084}$_{0.009}$ & \footnotesize{0.2350}$_{0.012}$ & \footnotesize{0.0598}$_{0.003}$ & \footnotesize{0.0592}$_{0.003}$ & \footnotesize{0.0522}$_{0.003}$ & \footnotesize{\bf 0.0315}$_{0.003}$ \\
    \footnotesize{DT} & \footnotesize{0.2265}$_{0.008}$ & \footnotesize{0.2200}$_{0.008}$ & \footnotesize{0.2081}$_{0.006}$ & \footnotesize{0.2141}$_{0.007}$ & \footnotesize{0.0595}$_{0.003}$ & \footnotesize{0.0589}$_{0.003}$ & \footnotesize{0.0546}$_{0.004}$ & \footnotesize{\bf 0.0354}$_{0.003}$ \\
    \footnotesize{SVM} & \footnotesize{0.1034}$_{0.004}$ & \footnotesize{0.1040}$_{0.005}$ & \footnotesize{0.1038}$_{0.005}$ & \footnotesize{0.1070}$_{0.005}$ & \footnotesize{0.0592}$_{0.003}$ & \footnotesize{0.0586}$_{0.003}$ & \footnotesize{0.0549}$_{0.003}$ & \footnotesize{\bf 0.0428}$_{0.002}$ \\
    \footnotesize{RF} & \footnotesize{0.1021}$_{0.002}$ & \footnotesize{0.0524}$_{0.004}$ & \footnotesize{0.0537}$_{0.003}$ & \footnotesize{0.0552}$_{0.003}$ & \footnotesize{0.0597}$_{0.003}$ & \footnotesize{0.0591}$_{0.003}$ & \footnotesize{0.0455}$_{0.003}$ & \footnotesize{\bf 0.0344}$_{0.003}$ \\
    \footnotesize{RBFSVM} & \footnotesize{0.0756}$_{0.003}$ & \footnotesize{0.0742}$_{0.003}$ & \footnotesize{0.0689}$_{0.003}$ & \footnotesize{0.0706}$_{0.003}$ & \footnotesize{0.0587}$_{0.003}$ & \footnotesize{0.0582}$_{0.003}$ & \footnotesize{0.0442}$_{0.004}$ & \footnotesize{\bf 0.0380}$_{0.004}$ \\
    \footnotesize{GBT} & \footnotesize{0.0615}$_{0.005}$ & \footnotesize{0.0649}$_{0.005}$ & \footnotesize{0.0593}$_{0.005}$ & \footnotesize{0.0590}$_{0.004}$ & \footnotesize{0.0587}$_{0.003}$ & \footnotesize{0.0581}$_{0.003}$ & \footnotesize{0.0464}$_{0.004}$ & \footnotesize{\bf 0.0393}$_{0.004}$ \\
  \hline
  \multicolumn{9}{c}{Accuracy}\\ \hline
    Model & Uncal. & Platt & Iso Reg & Hist bin & SBA-10 & SBAW-10 & SWC & SWC-HH \\ \hline
    \footnotesize{NB} & \footnotesize{0.8158}$_{0.011}$ & \footnotesize{0.8160}$_{0.011}$ & \footnotesize{0.8692}$_{0.007}$ & \footnotesize{0.8654}$_{0.007}$ & \footnotesize{0.9632}$_{0.003}$ & \footnotesize{0.9662}$_{0.002}$ & \footnotesize{0.9662}$_{0.003}$ & \footnotesize{\bf 0.9816}$_{0.001}$ \\
    \footnotesize{DT} & \footnotesize{0.8500}$_{0.009}$ & \footnotesize{0.8540}$_{0.008}$ & \footnotesize{0.8568}$_{0.006}$ & \footnotesize{0.8566}$_{0.005}$ & \footnotesize{0.9634}$_{0.003}$ & \footnotesize{0.9656}$_{0.002}$ & \footnotesize{0.9642}$_{0.002}$ & \footnotesize{\bf 0.9784}$_{0.002}$ \\
    \footnotesize{SVM} & \footnotesize{0.9328}$_{0.004}$ & \footnotesize{0.9340}$_{0.004}$ & \footnotesize{0.9330}$_{0.003}$ & \footnotesize{0.9328}$_{0.003}$ & \footnotesize{0.9638}$_{0.003}$ & \footnotesize{0.9660}$_{0.003}$ & \footnotesize{0.9628}$_{0.003}$ & \footnotesize{\bf 0.9744}$_{0.002}$ \\
    \footnotesize{RBFSVM} & \footnotesize{0.9454}$_{0.003}$ & \footnotesize{0.9454}$_{0.003}$ & \footnotesize{0.9544}$_{0.003}$ & \footnotesize{0.9524}$_{0.002}$ & \footnotesize{0.9640}$_{0.003}$ & \footnotesize{0.9662}$_{0.002}$ & \footnotesize{0.9694}$_{0.002}$ & \footnotesize{\bf 0.9788}$_{0.002}$ \\
    \footnotesize{GBT} & \footnotesize{0.9594}$_{0.002}$ & \footnotesize{0.9600}$_{0.003}$ & \footnotesize{0.9590}$_{0.003}$ & \footnotesize{0.9600}$_{0.003}$ & \footnotesize{0.9636}$_{0.003}$ & \footnotesize{0.9660}$_{0.003}$ & \footnotesize{0.9682}$_{0.002}$ & \footnotesize{\bf 0.9780}$_{0.002}$ \\
    \footnotesize{RF} & \footnotesize{0.9634}$_{0.002}$ & \footnotesize{0.9648}$_{0.002}$ & \footnotesize{0.9632}$_{0.002}$ & \footnotesize{0.9634}$_{0.002}$ & \footnotesize{0.9632}$_{0.003}$ & \footnotesize{0.9658}$_{0.003}$ & \footnotesize{0.9700}$_{0.002}$ & \footnotesize{\bf 0.9812}$_{0.002}$ \\
    \hline
  \end{tabular}
\end{table}

\newpage
\subsection{Multi-class tabular data}
\begin{figure}[ht]
  \centering
  \subfloat[mnist10 Brier score]{\includegraphics[width=3.25in]
    {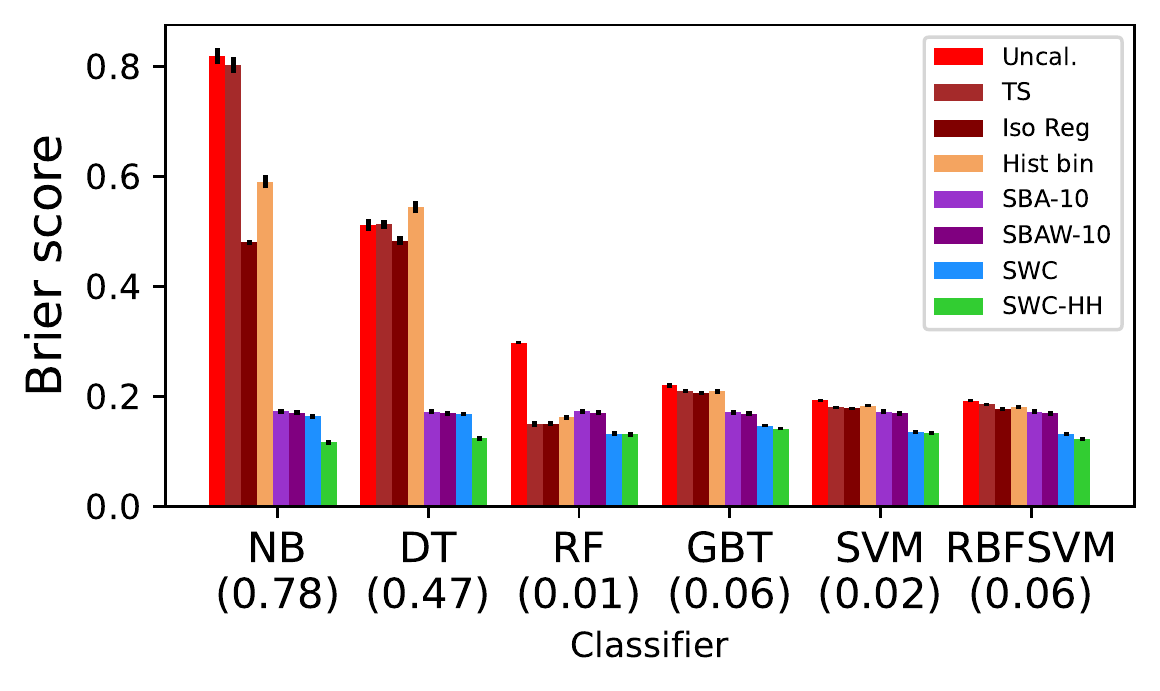}}
  \subfloat[mnist10 accuracy]{\includegraphics[width=3.25in]
    {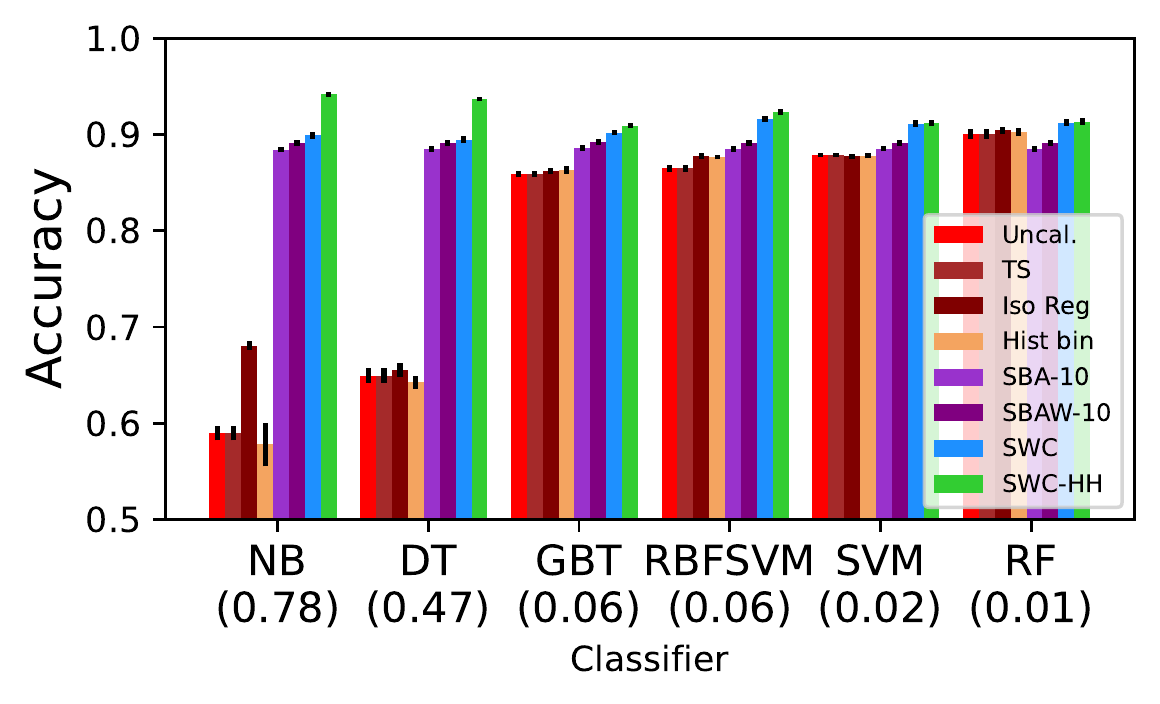}}
  \\
  \subfloat[fashion-mnist Brier score]{\includegraphics[width=3.25in]
    {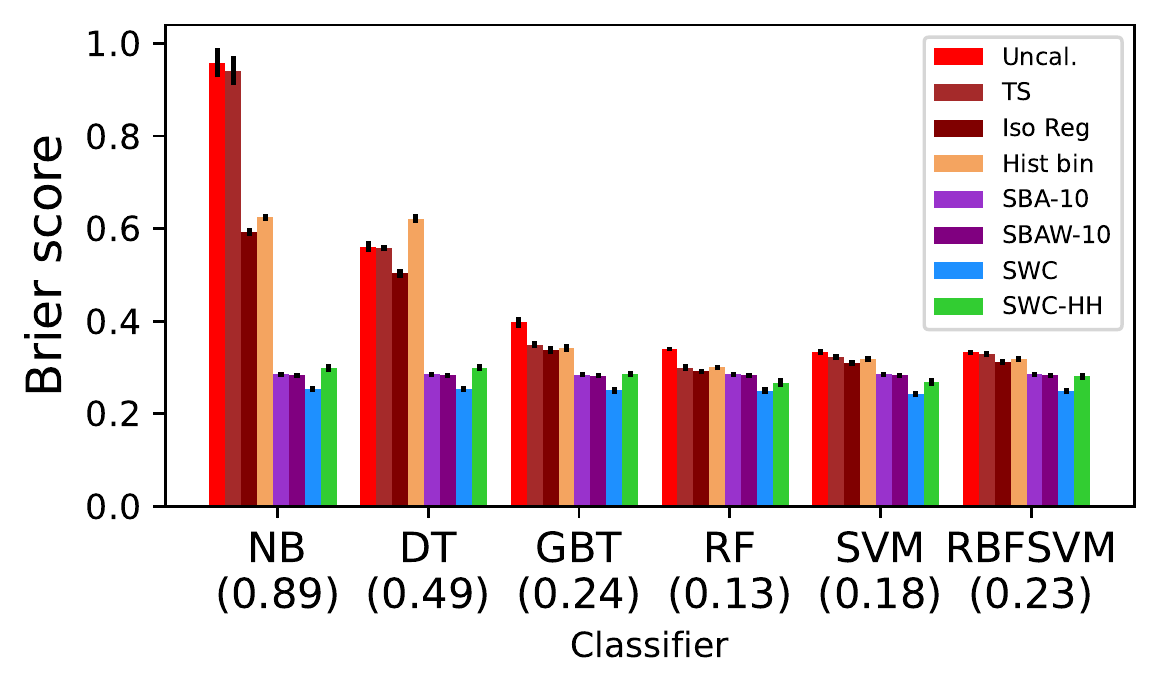}}
  \subfloat[fashion-mnist accuracy]{\includegraphics[width=3.25in]
    {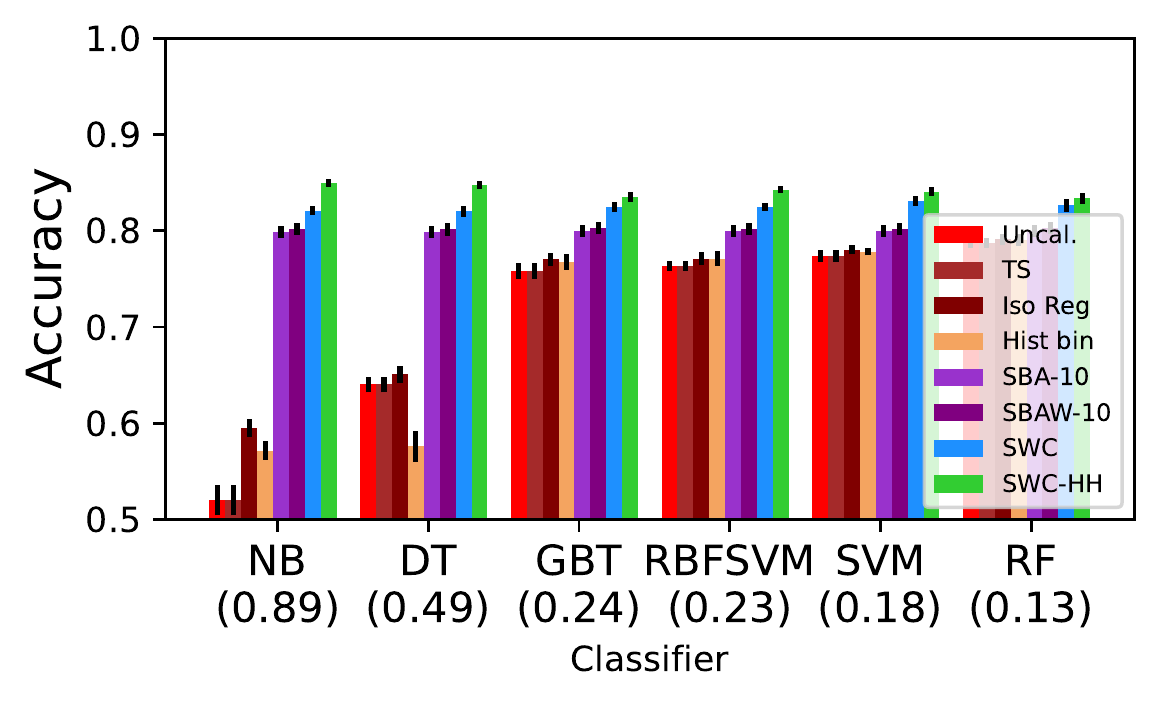}}
  \\
  \subfloat[letter Brier score]{\includegraphics[width=3.25in]
    {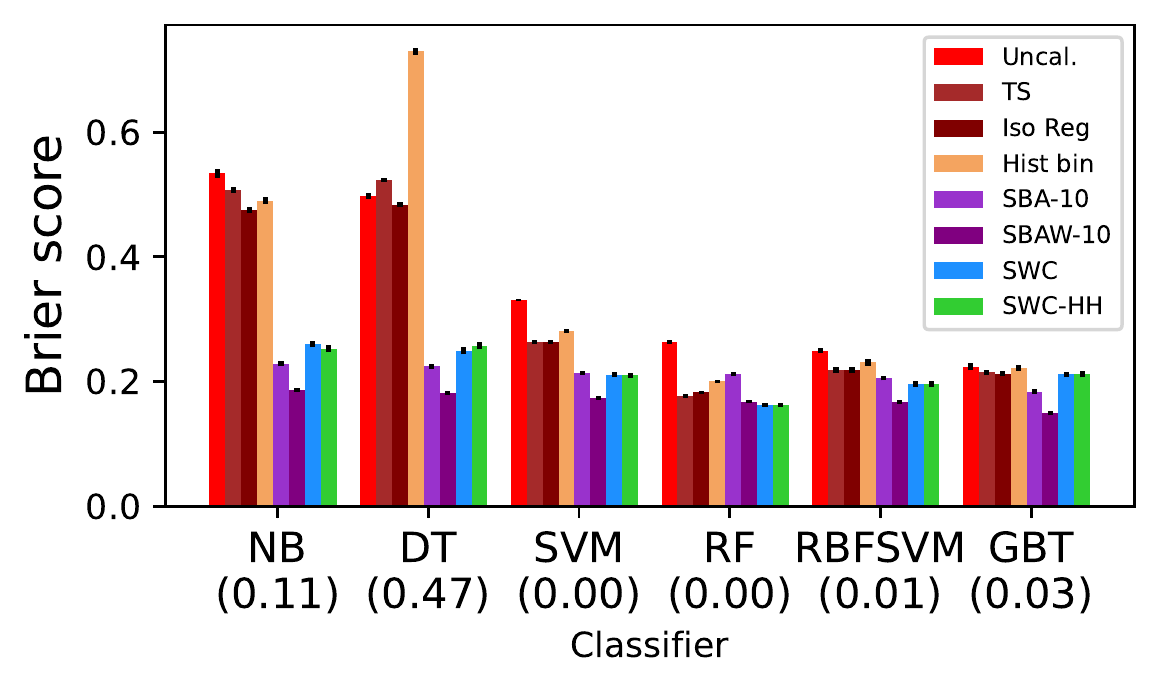}}
  \subfloat[letter accuracy]{\includegraphics[width=3.25in]
    {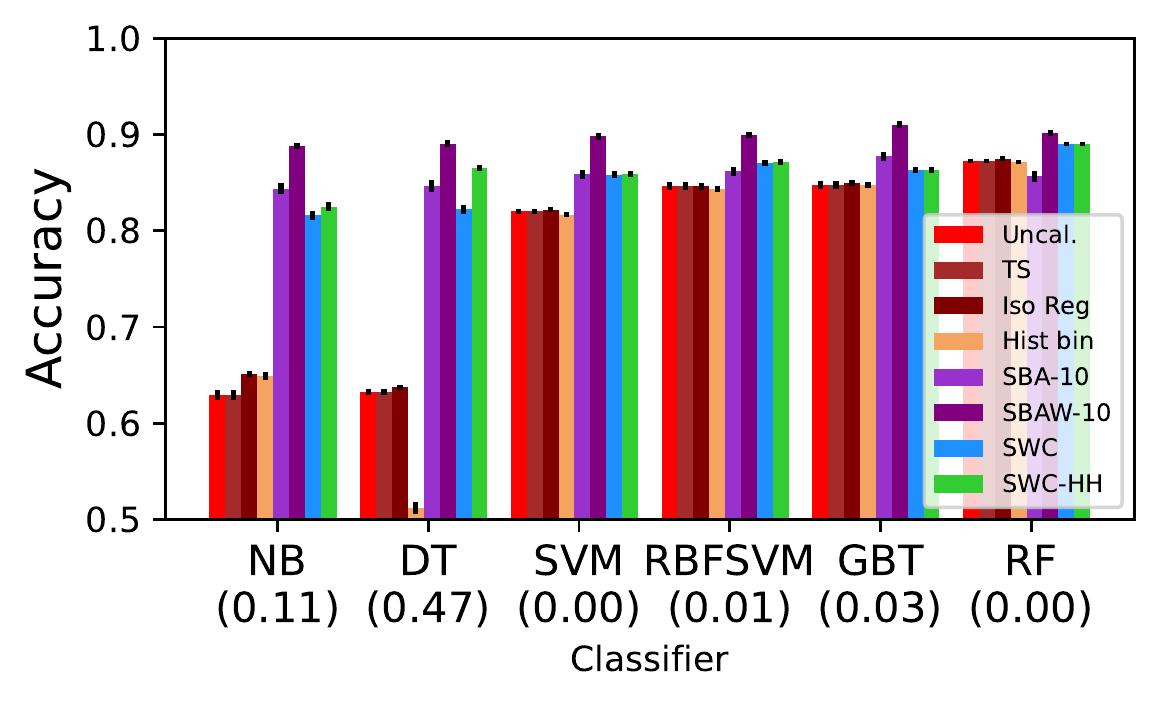}}
  \caption{Calibration performance (left) and accuracy (right)
    for the multi-class ``mnist10'', ``fashion-mnist'',
    and ''letter'' data sets
  (10 trials; error bars indicate one standard error). 
  Classifiers are sorted in order of improvement based on the
  uncalibrated classifier's score, and average HH values are below
  each classifier.}
    \label{fig:res-tabular-multi}
\end{figure}

Figure~\ref{fig:res-tabular-multi} and Tables~\ref{tab:mnist10}
to~\ref{tab:letter} present results for the
multi-class data sets ``mnist10'', ``fashion-mnist'', and ``letter'',
after \num{5000} calibration items were employed.
Temperature scaling in general only improved Brier score for
the tree-based methods (RF, GBT), and not consistently.
Histogram binning was again beneficial for the Naive Bayes models, but
it often made calibration worse for decision trees.
Isotonic regression yielded small additional improvements.

For ``mnist10'' and ``fashion-mnist'',
similarity-based calibration provided the best results.  SWC and
SWC-HH out-performed SBA-10.  SWC-HH usually improved
over SWC, except for the more challenging ``fashion-mnist'' data set.
In this data set, the filtering employed by SWC-HH (to ignore
calibration items with insufficient similarity) often resulted in no
calibration items remaining.  We handle this case by using the single
nearest neighbor, even if its similarity is below the threshold.  This
leads to values for $\hat{q}$ that are based only on one calibration
item.  In many cases, the single nearest neighbor belongs to the
correct class, yielding good accuracy, but when it is from an
incorrect class, the Brier score penalty is large.  However, the
SWC-HH results were still comparable or better than SBA-10 and the
global calibration methods on this data set.
In addition, SWC-HH yielded the best accuracy.

The ``letter'' data set is unusual in that SBAW-10 achieved the best
results (Figure~\ref{fig:res-tabular-multi}(e,f) and
Table~\ref{tab:letter}, except for the random forest classifier, which
is best calibrated using SWC or SWC-HH.  SBAW-10 on this data set also
outperforms the unweighted SBA-10 approach described
by~\cite{bella:sba09}.  We interpret this to mean that ``letter''
exhibits even stronger subpopulation locality than the others we have
studied.  These results reinforce the importance of employing some
form of weighting when calibrating using similarity.  However, the
choice of 10 neighbors to use does not always work best, and the ideal
constant would be difficult to estimate in advance. Therefore, we
recommend the use of the entire data set (via SWC or SWC-HH) as a more
robust solution. 

\begin{table}[ht]
  \caption{Results for mnist10
    ($n_{cal}=5000$, 10 trials).
    The best result(s) for each model (within 1 standard error, shown as a subscript) are in bold.}
  \label{tab:mnist10}
  \small
  \begin{tabular}{l|c|ccc|cccc} 
  \hline
  \multicolumn{9}{c}{Brier score}\\ \hline
    Model & Uncal. & TS & Iso Reg & Hist bin & SBA-10 & SBAW-10 & SWC & SWC-HH \\ \hline
    \footnotesize{NB} & \footnotesize{0.8193}$_{0.015}$ & \footnotesize{0.8031}$_{0.015}$ & \footnotesize{0.4802}$_{0.005}$ & \footnotesize{0.5910}$_{0.013}$ & \footnotesize{0.1729}$_{0.005}$ & \footnotesize{0.1702}$_{0.005}$ & \footnotesize{0.1637}$_{0.004}$ & \footnotesize{\bf 0.1164}$_{0.004}$ \\
    \footnotesize{DT} & \footnotesize{0.5119}$_{0.010}$ & \footnotesize{0.5128}$_{0.008}$ & \footnotesize{0.4832}$_{0.008}$ & \footnotesize{0.5442}$_{0.011}$ & \footnotesize{0.1722}$_{0.005}$ & \footnotesize{0.1695}$_{0.005}$ & \footnotesize{0.1678}$_{0.004}$ & \footnotesize{\bf 0.1236}$_{0.004}$ \\
    \footnotesize{RF} & \footnotesize{0.2977}$_{0.003}$ & \footnotesize{0.1495}$_{0.006}$ & \footnotesize{0.1505}$_{0.004}$ & \footnotesize{0.1620}$_{0.005}$ & \footnotesize{0.1726}$_{0.005}$ & \footnotesize{0.1699}$_{0.005}$ & \footnotesize{\bf 0.1323}$_{0.005}$ & \footnotesize{\bf 0.1311}$_{0.005}$ \\
    \footnotesize{GBT} & \footnotesize{0.2199}$_{0.005}$ & \footnotesize{0.2102}$_{0.004}$ & \footnotesize{0.2061}$_{0.004}$ & \footnotesize{0.2090}$_{0.004}$ & \footnotesize{0.1713}$_{0.005}$ & \footnotesize{0.1686}$_{0.005}$ & \footnotesize{0.1469}$_{0.003}$ & \footnotesize{\bf 0.1411}$_{0.003}$ \\
    \footnotesize{SVM} & \footnotesize{0.1926}$_{0.003}$ & \footnotesize{0.1805}$_{0.003}$ & \footnotesize{0.1786}$_{0.003}$ & \footnotesize{0.1830}$_{0.003}$ & \footnotesize{0.1719}$_{0.005}$ & \footnotesize{0.1692}$_{0.005}$ & \footnotesize{\bf 0.1347}$_{0.004}$ & \footnotesize{\bf 0.1338}$_{0.004}$ \\
    \footnotesize{RBFSVM} & \footnotesize{0.1919}$_{0.003}$ & \footnotesize{0.1854}$_{0.003}$ & \footnotesize{0.1766}$_{0.003}$ & \footnotesize{0.1813}$_{0.004}$ & \footnotesize{0.1717}$_{0.005}$ & \footnotesize{0.1690}$_{0.005}$ & \footnotesize{0.1312}$_{0.004}$ & \footnotesize{\bf 0.1223}$_{0.004}$ \\
  \hline
  \multicolumn{9}{c}{Accuracy}\\ \hline
    Model & Uncal. & TS & Iso Reg & Hist bin & SBA-10 & SBAW-10 & SWC & SWC-HH \\ \hline
    \footnotesize{NB} & \footnotesize{0.5895}$_{0.007}$ & \footnotesize{0.5895}$_{0.007}$ & \footnotesize{0.6806}$_{0.004}$ & \footnotesize{0.5779}$_{0.023}$ & \footnotesize{0.8841}$_{0.003}$ & \footnotesize{0.8909}$_{0.003}$ & \footnotesize{0.8991}$_{0.003}$ & \footnotesize{\bf 0.9415}$_{0.002}$ \\
    \footnotesize{DT} & \footnotesize{0.6495}$_{0.008}$ & \footnotesize{0.6495}$_{0.008}$ & \footnotesize{0.6551}$_{0.007}$ & \footnotesize{0.6427}$_{0.007}$ & \footnotesize{0.8847}$_{0.003}$ & \footnotesize{0.8910}$_{0.003}$ & \footnotesize{0.8947}$_{0.004}$ & \footnotesize{\bf 0.9368}$_{0.002}$ \\
    \footnotesize{GBT} & \footnotesize{0.8589}$_{0.004}$ & \footnotesize{0.8589}$_{0.004}$ & \footnotesize{0.8618}$_{0.003}$ & \footnotesize{0.8631}$_{0.004}$ & \footnotesize{0.8856}$_{0.003}$ & \footnotesize{0.8918}$_{0.003}$ & \footnotesize{0.9018}$_{0.003}$ & \footnotesize{\bf 0.9091}$_{0.002}$ \\
    \footnotesize{RBFSVM} & \footnotesize{0.8647}$_{0.003}$ & \footnotesize{0.8647}$_{0.003}$ & \footnotesize{0.8777}$_{0.003}$ & \footnotesize{0.8765}$_{0.002}$ & \footnotesize{0.8852}$_{0.003}$ & \footnotesize{0.8914}$_{0.003}$ & \footnotesize{0.9159}$_{0.003}$ & \footnotesize{\bf 0.9230}$_{0.003}$ \\
    \footnotesize{SVM} & \footnotesize{0.8785}$_{0.002}$ & \footnotesize{0.8785}$_{0.002}$ & \footnotesize{0.8772}$_{0.002}$ & \footnotesize{0.8780}$_{0.003}$ & \footnotesize{0.8853}$_{0.003}$ & \footnotesize{0.8912}$_{0.003}$ & \footnotesize{\bf 0.9112}$_{0.003}$ & \footnotesize{\bf 0.9118}$_{0.003}$ \\
    \footnotesize{RF} & \footnotesize{0.9006}$_{0.005}$ & \footnotesize{0.9006}$_{0.005}$ & \footnotesize{0.9043}$_{0.004}$ & \footnotesize{0.9023}$_{0.004}$ & \footnotesize{0.8849}$_{0.003}$ & \footnotesize{0.8908}$_{0.003}$ & \footnotesize{\bf 0.9123}$_{0.003}$ & \footnotesize{\bf 0.9134}$_{0.003}$ \\
    \hline
  \end{tabular}
\end{table}

\begin{table}[ht]
  \caption{Results for fashion-mnist
    ($n_{cal}=5000$, 10 trials).
    The best result(s) for each model (within 1 standard error, shown as a subscript) are in bold.}
  \label{tab:fashion-mnist}
  \small
  \begin{tabular}{l|c|ccc|cccc} 
  \hline
  \multicolumn{9}{c}{Brier score}\\ \hline
    Model & Uncal. & TS & Iso Reg & Hist bin & SBA-10 & SBAW-10 & SWC & SWC-HH \\ \hline
    \footnotesize{NB} & \footnotesize{0.9586}$_{0.032}$ & \footnotesize{0.9415}$_{0.032}$ & \footnotesize{0.5926}$_{0.008}$ & \footnotesize{0.6247}$_{0.008}$ & \footnotesize{0.2857}$_{0.005}$ & \footnotesize{0.2833}$_{0.005}$ & \footnotesize{\bf 0.2531}$_{0.006}$ & \footnotesize{0.2991}$_{0.008}$ \\
    \footnotesize{DT} & \footnotesize{0.5613}$_{0.011}$ & \footnotesize{0.5579}$_{0.007}$ & \footnotesize{0.5032}$_{0.009}$ & \footnotesize{0.6216}$_{0.009}$ & \footnotesize{0.2855}$_{0.005}$ & \footnotesize{0.2832}$_{0.005}$ & \footnotesize{\bf 0.2536}$_{0.007}$ & \footnotesize{0.2996}$_{0.007}$ \\
    \footnotesize{GBT} & \footnotesize{0.3976}$_{0.012}$ & \footnotesize{0.3493}$_{0.008}$ & \footnotesize{0.3370}$_{0.008}$ & \footnotesize{0.3427}$_{0.009}$ & \footnotesize{0.2845}$_{0.005}$ & \footnotesize{0.2821}$_{0.005}$ & \footnotesize{\bf 0.2504}$_{0.007}$ & \footnotesize{0.2859}$_{0.006}$ \\
    \footnotesize{RF} & \footnotesize{0.3400}$_{0.004}$ & \footnotesize{0.2997}$_{0.007}$ & \footnotesize{0.2913}$_{0.006}$ & \footnotesize{0.3000}$_{0.006}$ & \footnotesize{0.2855}$_{0.005}$ & \footnotesize{0.2832}$_{0.005}$ & \footnotesize{\bf 0.2497}$_{0.007}$ & \footnotesize{0.2674}$_{0.009}$ \\
    \footnotesize{SVM} & \footnotesize{0.3333}$_{0.007}$ & \footnotesize{0.3226}$_{0.007}$ & \footnotesize{0.3101}$_{0.006}$ & \footnotesize{0.3176}$_{0.007}$ & \footnotesize{0.2856}$_{0.005}$ & \footnotesize{0.2832}$_{0.005}$ & \footnotesize{\bf 0.2428}$_{0.007}$ & \footnotesize{0.2682}$_{0.008}$ \\
    \footnotesize{RBFSVM} & \footnotesize{0.3327}$_{0.006}$ & \footnotesize{0.3290}$_{0.007}$ & \footnotesize{0.3120}$_{0.006}$ & \footnotesize{0.3182}$_{0.006}$ & \footnotesize{0.2855}$_{0.005}$ & \footnotesize{0.2832}$_{0.005}$ & \footnotesize{\bf 0.2486}$_{0.007}$ & \footnotesize{0.2807}$_{0.007}$ \\
  \hline
  \multicolumn{9}{c}{Accuracy}\\ \hline
    Model & Uncal. & TS & Iso Reg & Hist bin & SBA-10 & SBAW-10 & SWC & SWC-HH \\ \hline
    \footnotesize{NB} & \footnotesize{0.5202}$_{0.016}$ & \footnotesize{0.5202}$_{0.016}$ & \footnotesize{0.5950}$_{0.009}$ & \footnotesize{0.5716}$_{0.010}$ & \footnotesize{0.7986}$_{0.006}$ & \footnotesize{0.8022}$_{0.006}$ & \footnotesize{0.8206}$_{0.005}$ & \footnotesize{\bf 0.8498}$_{0.004}$ \\
    \footnotesize{DT} & \footnotesize{0.6402}$_{0.008}$ & \footnotesize{0.6402}$_{0.008}$ & \footnotesize{0.6510}$_{0.009}$ & \footnotesize{0.5760}$_{0.016}$ & \footnotesize{0.7988}$_{0.006}$ & \footnotesize{0.8014}$_{0.006}$ & \footnotesize{0.8200}$_{0.005}$ & \footnotesize{\bf 0.8476}$_{0.004}$ \\
    \footnotesize{GBT} & \footnotesize{0.7584}$_{0.008}$ & \footnotesize{0.7584}$_{0.008}$ & \footnotesize{0.7702}$_{0.007}$ & \footnotesize{0.7676}$_{0.008}$ & \footnotesize{0.7996}$_{0.006}$ & \footnotesize{0.8028}$_{0.006}$ & \footnotesize{0.8246}$_{0.005}$ & \footnotesize{\bf 0.8352}$_{0.005}$ \\
    \footnotesize{RBFSVM} & \footnotesize{0.7634}$_{0.005}$ & \footnotesize{0.7634}$_{0.005}$ & \footnotesize{0.7710}$_{0.007}$ & \footnotesize{0.7708}$_{0.008}$ & \footnotesize{0.7998}$_{0.006}$ & \footnotesize{0.8022}$_{0.006}$ & \footnotesize{0.8244}$_{0.004}$ & \footnotesize{\bf 0.8426}$_{0.004}$ \\
    \footnotesize{SVM} & \footnotesize{0.7738}$_{0.006}$ & \footnotesize{0.7738}$_{0.006}$ & \footnotesize{0.7804}$_{0.004}$ & \footnotesize{0.7782}$_{0.004}$ & \footnotesize{0.7996}$_{0.006}$ & \footnotesize{0.8022}$_{0.006}$ & \footnotesize{0.8312}$_{0.005}$ & \footnotesize{\bf 0.8406}$_{0.004}$ \\
    \footnotesize{RF} & \footnotesize{0.7872}$_{0.005}$ & \footnotesize{0.7872}$_{0.005}$ & \footnotesize{0.7912}$_{0.006}$ & \footnotesize{0.7902}$_{0.006}$ & \footnotesize{0.7996}$_{0.006}$ & \footnotesize{0.8024}$_{0.006}$ & \footnotesize{0.8262}$_{0.006}$ & \footnotesize{\bf 0.8336}$_{0.006}$ \\
    \hline
  \end{tabular}
\end{table}

\begin{table}[ht]
  \caption{Results for letter
    ($n_{cal}=5000$, 10 trials).
    The best result(s) for each model (within 1 standard error, shown as a subscript) are in bold.}
  \label{tab:letter}
  \small
  \begin{tabular}{l|c|ccc|cccc} 
  \hline
  \multicolumn{9}{c}{Brier score}\\ \hline
    Model & Uncal. & TS & Iso Reg & Hist bin & SBA-10 & SBAW-10 & SWC & SWC-HH \\ \hline
    \footnotesize{NB} & \footnotesize{0.5342}$_{0.007}$ & \footnotesize{0.5073}$_{0.005}$ & \footnotesize{0.4753}$_{0.005}$ & \footnotesize{0.4899}$_{0.005}$ & \footnotesize{0.2284}$_{0.004}$ & \footnotesize{\bf 0.1863}$_{0.003}$ & \footnotesize{0.2603}$_{0.005}$ & \footnotesize{0.2527}$_{0.006}$ \\
    \footnotesize{DT} & \footnotesize{0.4975}$_{0.004}$ & \footnotesize{0.5237}$_{0.003}$ & \footnotesize{0.4839}$_{0.004}$ & \footnotesize{0.7296}$_{0.006}$ & \footnotesize{0.2243}$_{0.004}$ & \footnotesize{\bf 0.1812}$_{0.003}$ & \footnotesize{0.2492}$_{0.006}$ & \footnotesize{0.2576}$_{0.006}$ \\
    \footnotesize{SVM} & \footnotesize{0.3311}$_{0.002}$ & \footnotesize{0.2629}$_{0.003}$ & \footnotesize{0.2632}$_{0.003}$ & \footnotesize{0.2806}$_{0.003}$ & \footnotesize{0.2141}$_{0.003}$ & \footnotesize{\bf 0.1737}$_{0.003}$ & \footnotesize{0.2107}$_{0.004}$ & \footnotesize{0.2102}$_{0.004}$ \\
    \footnotesize{RF} & \footnotesize{0.2633}$_{0.003}$ & \footnotesize{0.1769}$_{0.003}$ & \footnotesize{0.1830}$_{0.003}$ & \footnotesize{0.2002}$_{0.003}$ & \footnotesize{0.2121}$_{0.004}$ & \footnotesize{0.1679}$_{0.003}$ & \footnotesize{\bf 0.1630}$_{0.003}$ & \footnotesize{\bf 0.1630}$_{0.003}$ \\
    \footnotesize{RBFSVM} & \footnotesize{0.2495}$_{0.004}$ & \footnotesize{0.2186}$_{0.005}$ & \footnotesize{0.2189}$_{0.005}$ & \footnotesize{0.2308}$_{0.005}$ & \footnotesize{0.2056}$_{0.004}$ & \footnotesize{\bf 0.1673}$_{0.003}$ & \footnotesize{0.1963}$_{0.005}$ & \footnotesize{0.1957}$_{0.005}$ \\
    \footnotesize{GBT} & \footnotesize{0.2237}$_{0.006}$ & \footnotesize{0.2147}$_{0.005}$ & \footnotesize{0.2129}$_{0.004}$ & \footnotesize{0.2219}$_{0.004}$ & \footnotesize{0.1835}$_{0.004}$ & \footnotesize{\bf 0.1491}$_{0.003}$ & \footnotesize{0.2114}$_{0.005}$ & \footnotesize{0.2115}$_{0.005}$ \\
  \hline
  \multicolumn{9}{c}{Accuracy}\\ \hline
    Model & Uncal. & TS & Iso Reg & Hist bin & SBA-10 & SBAW-10 & SWC & SWC-HH \\ \hline
    \footnotesize{NB} & \footnotesize{0.6291}$_{0.005}$ & \footnotesize{0.6291}$_{0.005}$ & \footnotesize{0.6511}$_{0.003}$ & \footnotesize{0.6492}$_{0.004}$ & \footnotesize{0.8438}$_{0.006}$ & \footnotesize{\bf 0.8881}$_{0.003}$ & \footnotesize{0.8160}$_{0.005}$ & \footnotesize{0.8251}$_{0.004}$ \\
    \footnotesize{DT} & \footnotesize{0.6321}$_{0.003}$ & \footnotesize{0.6321}$_{0.003}$ & \footnotesize{0.6375}$_{0.003}$ & \footnotesize{0.5119}$_{0.006}$ & \footnotesize{0.8465}$_{0.006}$ & \footnotesize{\bf 0.8904}$_{0.003}$ & \footnotesize{0.8223}$_{0.005}$ & \footnotesize{0.8654}$_{0.003}$ \\
    \footnotesize{SVM} & \footnotesize{0.8202}$_{0.002}$ & \footnotesize{0.8202}$_{0.002}$ & \footnotesize{0.8217}$_{0.003}$ & \footnotesize{0.8168}$_{0.003}$ & \footnotesize{0.8584}$_{0.005}$ & \footnotesize{\bf 0.8982}$_{0.004}$ & \footnotesize{0.8581}$_{0.003}$ & \footnotesize{0.8589}$_{0.003}$ \\
    \footnotesize{RBFSVM} & \footnotesize{0.8461}$_{0.004}$ & \footnotesize{0.8461}$_{0.004}$ & \footnotesize{0.8459}$_{0.004}$ & \footnotesize{0.8434}$_{0.003}$ & \footnotesize{0.8615}$_{0.005}$ & \footnotesize{\bf 0.8998}$_{0.003}$ & \footnotesize{0.8701}$_{0.003}$ & \footnotesize{0.8711}$_{0.003}$ \\
    \footnotesize{GBT} & \footnotesize{0.8479}$_{0.004}$ & \footnotesize{0.8479}$_{0.004}$ & \footnotesize{0.8492}$_{0.003}$ & \footnotesize{0.8474}$_{0.003}$ & \footnotesize{0.8774}$_{0.004}$ & \footnotesize{\bf 0.9103}$_{0.003}$ & \footnotesize{0.8629}$_{0.003}$ & \footnotesize{0.8633}$_{0.003}$ \\
    \footnotesize{RF} & \footnotesize{0.8722}$_{0.002}$ & \footnotesize{0.8722}$_{0.002}$ & \footnotesize{0.8746}$_{0.002}$ & \footnotesize{0.8710}$_{0.002}$ & \footnotesize{0.8565}$_{0.006}$ & \footnotesize{\bf 0.9013}$_{0.003}$ & \footnotesize{0.8904}$_{0.002}$ & \footnotesize{0.8904}$_{0.002}$ \\
    \hline
  \end{tabular}
\end{table}

\end{document}